\documentclass[aip, pof, preprint, numerical]{revtex4-1}

\usepackage{graphicx}% Include figure files
\usepackage{dcolumn}% Align table columns on decimal point
\usepackage{bm}% bold math
\usepackage[mathlines]{lineno}% Enable numbering of text and display math
\usepackage[utf8]{inputenc}
\usepackage[T1]{fontenc}
\usepackage{mathptmx}
\usepackage{etoolbox}
\usepackage{amsmath}
\usepackage{graphicx}       % required for inserting images
\usepackage[utf8]{inputenc} % allow utf-8 input
\usepackage[T1]{fontenc}    % use 8-bit T1 fonts
\usepackage{hyperref}       % hyperlinks
\usepackage{url}            % simple URL typesetting
\usepackage{booktabs}       % professional-quality tables
\usepackage{amsfonts}       % blackboard math symbols
\usepackage{nicefrac}       % compact symbols for 1/2, etc.
\usepackage{microtype}      % microtypography
\usepackage{xcolor}         % colors
\usepackage{soul}
\usepackage{mathtools}
\usepackage{epstopdf}
\usepackage[noend]{algpseudocode}
\usepackage{algorithm}

\newcommand\norm[1]{\left\lVert#1\right\rVert}

\makeatletter
\def\@email#1#2{%
 \endgroup
 \patchcmd{\titleblock@produce}
  {\frontmatter@RRAPformat}
  {\frontmatter@RRAPformat{\produce@RRAP{*#1\href{mailto:#2}{#2}}}\frontmatter@RRAPformat}
  {}{}
}%
\makeatother
\begin{document}

\preprint{AIP/123-QED}

\title[Transient-CoMLSim]{A domain decomposition-based autoregressive deep learning model for unsteady and nonlinear partial differential equations}

\author{S. Nidhan}
\email{sheel.nidhan@ansys.com}
\affiliation{Ansys, Inc., San Jose, CA}%

\author{H. Jiang}%
\affiliation{Ansys, Inc., San Jose, CA}%

\author{L. Ghule}
\affiliation{Ansys, Inc., Canonsburg, PA}%

\author{C. Umphrey}
\affiliation{Ansys, Inc., Salt Lake City, UT}%

\author{R. Ranade\footnote{R. Ranade contributed to this work during his tenure at Ansys, Inc.}}
\affiliation{NVIDIA, Santa Clara, CA}%

\author{J. Pathak}
\affiliation{Ansys, Inc., San Jose, CA}%

% \date{\today}

\begin{abstract}
In this paper, we propose a domain-decomposition-based deep learning (DL) framework, named transient-CoMLSim, to accurately model unstable and nonlinear partial differential equations (PDEs). The framework consists of two key components: (a) a convolutional neural network (CNN)-based autoencoder architecture and (b) an autoregressive model composed of fully connected layers. Unlike existing state-of-the-art methods that operate on the entire computational domain, our CNN-based autoencoder computes a lower-dimensional basis for solution and condition fields represented on subdomains. Timestepping is performed entirely in the latent space, generating embeddings of the solution variables from the time history of embeddings of solution and condition variables. This approach not only reduces computational complexity, but also enhances scalability, making it well-suited for large-scale simulations. Furthermore, to improve the stability of our rollouts, we employ a curriculum learning (CL) approach during the training of the autoregressive model. The domain-decomposition strategy enables scaling to out-of-distribution domain sizes while maintaining the accuracy of predictions -- a feature not easily integrated into popular DL-based approaches for physics simulations. We benchmark our model against two widely used DL architectures, Fourier Neural Operator (FNO) and U-Net, and demonstrate that our framework outperforms them in terms of accuracy, extrapolation to unseen timesteps, and stability for a wide range of use cases.
\end{abstract}
\maketitle

\section{Introduction}
In recent years, the intersection of deep learning (ML) and physics-based modeling has gained significant attention, fueled by the potential of data-driven techniques to enhance our understanding and prediction of complex physical phenomena. Several methods, both data-driven and physics-informed\cite{raissi2018hidden,li2020fourier,li2023fourier,lu2021learning,pfaff2020learning,sanchez2020learning,stachenfeld2021learned}, have been developed to create surrogates to solve partial differential equations (PDEs). These deep learning (DL) models offer the promise of reducing the computational time required to solve highly nonlinear PDEs. In addition to developing DL models driven purely by data, there have been efforts to create hybrid models that combine the strengths of traditional PDE solvers with the capabilities of DL methods \citep{ranade2022composable,ranade2022thermal,ranade2021discretizationnet,he2023solving,zhang2022hybrid,kahana2023geometry}. These hybrid approaches leverage the generalization ability of PDE solvers while using DL techniques to effectively model high-dimensional manifolds by identifying repeated patterns in the solution and geometry fields. As a result, they serve as an augmentation of traditional PDE solvers\citep{bar2019learning,kochkov2021machine,um2020solver,tathawadekar2021hybrid,brahmachary2024unsteady,ramos2022control,list2022learned,coros2021differentiable,holl2020learning}. These DL-based models have demonstrated significant potential in fields where PDE modeling is essential, such as weather prediction \citep{gupta2022towards,lam2022graphcast}, 3-D aerodynamics predictions \cite{li2024geometry}, etc.

Among the various frameworks developed for physics-based problems, physics-informed neural networks (PINNs) \cite{raissi2018hidden} and Fourier neural operators (FNOs) \cite{li2020fourier,li2024geometry} are two of the most prevalent, both with distinct strengths and applications. PINNs leverage the underlying PDEs by incorporating the residual governing equations directly into the loss function during training. This approach allows them to solve both forward and inverse problems while respecting physical laws, making them particularly useful in scenarios where data is scarce but the physics is well understood. In contrast, FNOs take a different approach by learning operators in the frequency domain. Using Fourier transforms, FNOs capture global information and efficiently handle complex, high-dimensional problems.  In recent years, graph-based approaches using graph neural networks (GNNs) have also seen significant growth, as demonstrated by the works of Brandstetter et al. \cite{brandstetter2022message}, Xu et al. \cite{xu2021conditionally}, and Boussif et al. \cite{boussif2022magnet}. One of the key advantages of GNN-based methods is the seamless formulation of unstructured simulations, prevalent in PDE modeling of complex physics \cite{mavriplis1997unstructured}, on graphs.

One of the key distinctions of existing deep learning methods, such as those mentioned above, is that most of them are global in nature. By `global', we mean that they learn the solution distribution across the entire computational domain simultaneously. This contrasts with traditional PDE solvers, which derive solutions at computational grid points based on local differentials. Due to their global learning approach, these deep learning models often struggle to accurately resolving small-scale features or extrapolating to out-of-distribution domain sizes. Moreover, they face additional challenges such as the inability to represent complex geometries, difficulty in scaling to high-dimensional problems with the high-resolution meshes that are quite common in industrial simulations, exorbitantly large training times, prohibitive GPU memory requirements, and limited generalization to out-of-distribution input spaces. These issues significantly hinder the practical application of deep learning in real-world problems. Inspired by the local approach of traditional PDE solvers, Ranade et al. \cite{ranade2022composable} introduced the Composable Machine Learning Simulator (CoMLSim) framework for steady-state PDEs. Unlike global deep learning methods, CoMLSim focuses on learning solution distributions and conditions in low-dimensional spaces within local subdomains rather than across the entire computational domain. It then stitches these local solutions together by ensuring consistency and continuity across subdomains using flux schemes learned with neural networks. This approach integrates two key features: (a) a local-learning strategy and (b) a low-dimensional learning. Ranade et al. demonstrated that CoMLSim outperformed DL methods learning on the global level, such as FNO \cite{li2020fourier}, DeepONet \cite{lu2021learning}, and U-Net\cite{ronneberger2015u} on a range of steady-state PDEs, from simple Poisson's equations to industrial applications like 3D Reynolds Averaged Navier-Stokes (RANS) equations and chip cooling\cite{ranade2022thermal}. 

In this work, we extend the CoMLSim framework to handle transient PDEs, introducing what we call the transient-CoMLSim. The paper is organized as follows: Section \ref{sec:methodology} outlines the methodology behind the transient-CoMLSim approach. Section \ref{sec:results} presents the results, comparing transient-CoMLSim with benchmark models and evaluating its extrapolation capabilities across four different PDEs. Additionally, we include ablation studies for the time integrator network in Section \ref{sec:results}. The study is concluded in Section \ref{sec:conclusions}.

\section{Methodology}\label{sec:methodology}

\begin{figure}[htbp]
\centering
\includegraphics[trim={0.0cm, 0cm, 0.0cm, 0cm},clip=true,width=1\linewidth]{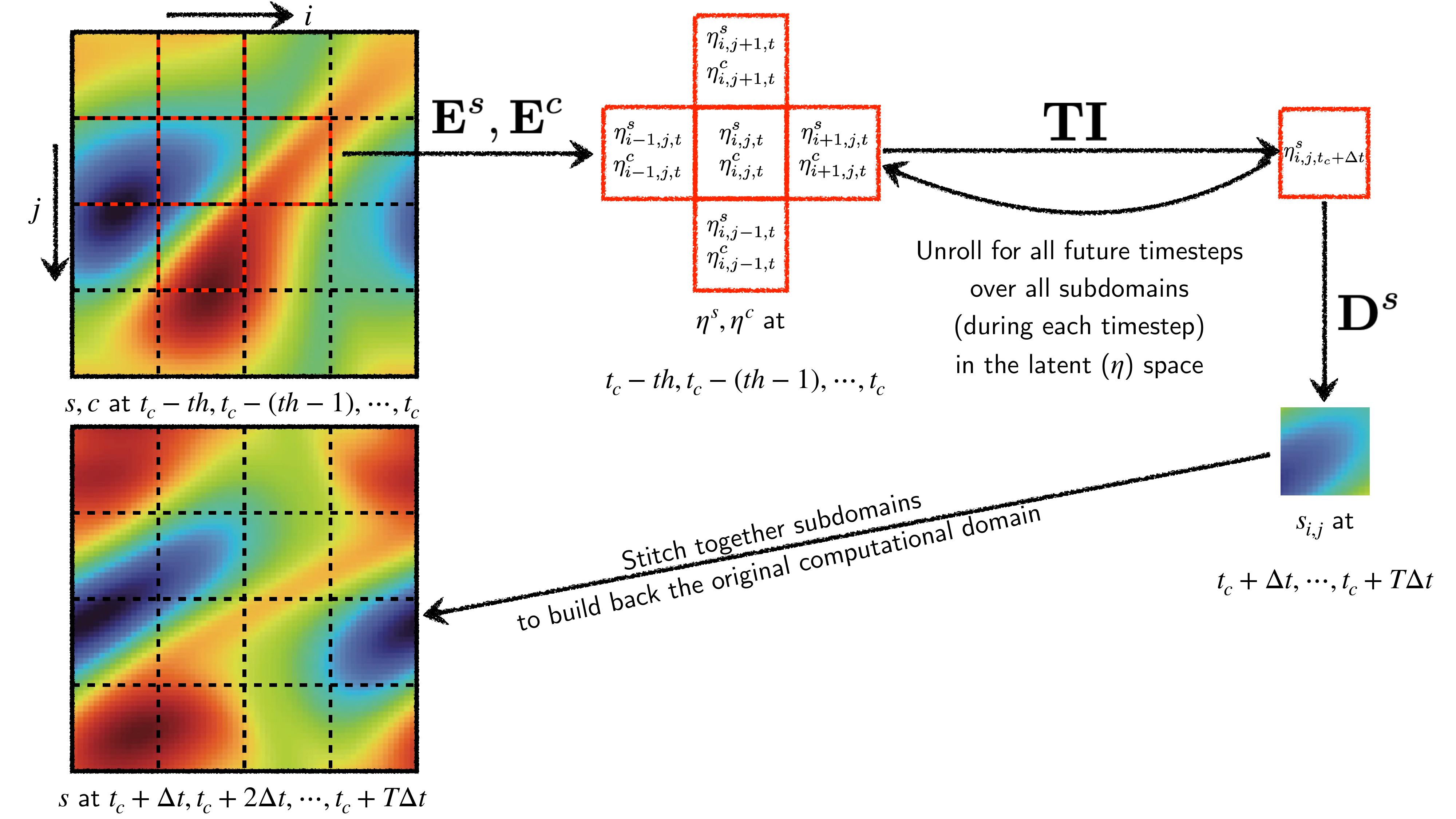}
\caption{Schematic of the transient-CoMLSim framework for a 2-D problem setup. $\mathbf{E}^s$ and $\mathbf{E}^c$ are solution and condition encoders. $\mathbf{D}^s$ refers to the solution decoder. $\mathbf{TI}$ is the time-integrator network. $\eta^s$ and $\eta^c$ correspond to solution and condition latent embeddings. Model predicts upto $T$ timesteps into the future. For 3-D problem setup, the setup can be easily extended to include 7 neighbors (left, right, top, bottom, front, back).}
\label{fig:schematic_overall}
\end{figure}

Let us assume a solution variable $s$ and a condition variable $c$ defined on a global computational domain $\Omega_G$. For example, in an unsteady heat diffusion equation with a moving laser source, the temperature $T$ serves as the solution variable, while the laser power $Q$ acts as the condition variable. Our methodology entails first decomposing $\Omega$ into multiple smaller subdomains $\Omega_1, \Omega_2, \cdots, \Omega_N$ such that $\Omega_1 \cup \Omega_2 \cdots \cup \Omega_N = \Omega_G$. Thereafter, solution and condition on each of those subdomains are encoded using a convolutional neural network (CNN) based autoencoder, named $\mathbf{AE}$ hereafter, to obtain latent encodings $\{\eta_{1}^{s}, \eta_{2}^{s}, \cdots, \eta_{N}^{s}\}$ and $\{\eta_1^{c}, \eta_2^{c}, \cdots, \eta_N^{c}\}$ for the solution and the condition variable, respectively.

\begin{equation}
    \eta_n^{s} = \mathbf{E}^s(s_n), \eta_n^c  = \mathbf{E}^c(c_n),
    \label{encoder_eq}
\end{equation}

\begin{equation}
    \hat{s}_n = \mathbf{D}^s(\eta_n^{s}), \hat{c}_n  = \mathbf{D}^c(\eta_n^c).
    \label{decoder_eq}
\end{equation}

Here, $\mathbf{E}$ and $\mathbf{D}$ are the encoder and decoder components of our CNN-based autoencoder, i.e., $\hat{s}_n = \mathbf{AE}^s(s_n) = \mathbf{D}^s(\mathbf{E}^s(s_n))$ and $\hat{c}_n = \mathbf{AE}^c(c_n) = \mathbf{D}^c(\mathbf{E}^c(c_n))$. Here, $n$ denotes the $n^{th}$ subdomain. Our CNN-based autoencoder, described in detail in section \ref{sec:mesh_autoencoder}, encodes the subdomains into low-dimensional latent embeddings. Representing solutions in a non-linear lower-dimensional basis offers benefits such as reduced computational costs and memory usage for time integration, as well as an efficient and dense representation of solution and condition fields, that enables accurate learning. Thereafter, the time integrator network, named $\mathbf{TI}$ hereafter, learns a robust mapping to the future timestep in the latent space for a given subdomain from: (a) the history of previous timesteps of solution and condition and (b) information of the surrounding subdomains:

\begin{equation}
    \eta^s_{n, t_c+\Delta t} = \mathbf{TI}(\eta^s_{\mathcal{I}, \mathcal{T}_s}, \eta^c_{\mathcal{I}, \mathcal{T}_c}; \theta). 
    \label{time_int_eq}
\end{equation}

Here, $\mathcal{I}$ denotes the collection of subdomains in the neighborhood of the subdomain $n$ and $\mathcal{T}$ denotes the time history (consisting of previous $th$ timesteps) of all the previous embeddings for any subdomain in the vicinity of $n$ (including $n$ itself), i.e., $\mathcal{T}_s \equiv (t_c, t_c-\Delta t, \cdots, t_c-th\Delta t)$ and $\mathcal{T}_c \equiv (t_c, t_c-\Delta t, \cdots, t_c-th\Delta t)$. For PDEs without any time varying condition field, one can only consider the solution encodings ($\eta^s$) in the equation \ref{time_int_eq}. Likewise, multiple solution and condition fields can be handled by including their respective $\eta$
in the input (for solution and condition fields) and the output (for solution fields only). A broad overview of the model is depicted in figure \ref{fig:schematic_overall}. In the subsequent subsections, we will delve deeper into the details of the CNN-based autoencoder $\mathbf{AE}$ and time integrator $\mathbf{TI}$.

\subsection{Local learning on a structured computational domain} \label{sec:mesh_autoencoder}

\begin{figure}[htbp]
\centering
\includegraphics[trim={0.0cm, 10cm, 0.0cm, 0cm},clip=true,width=1\linewidth]{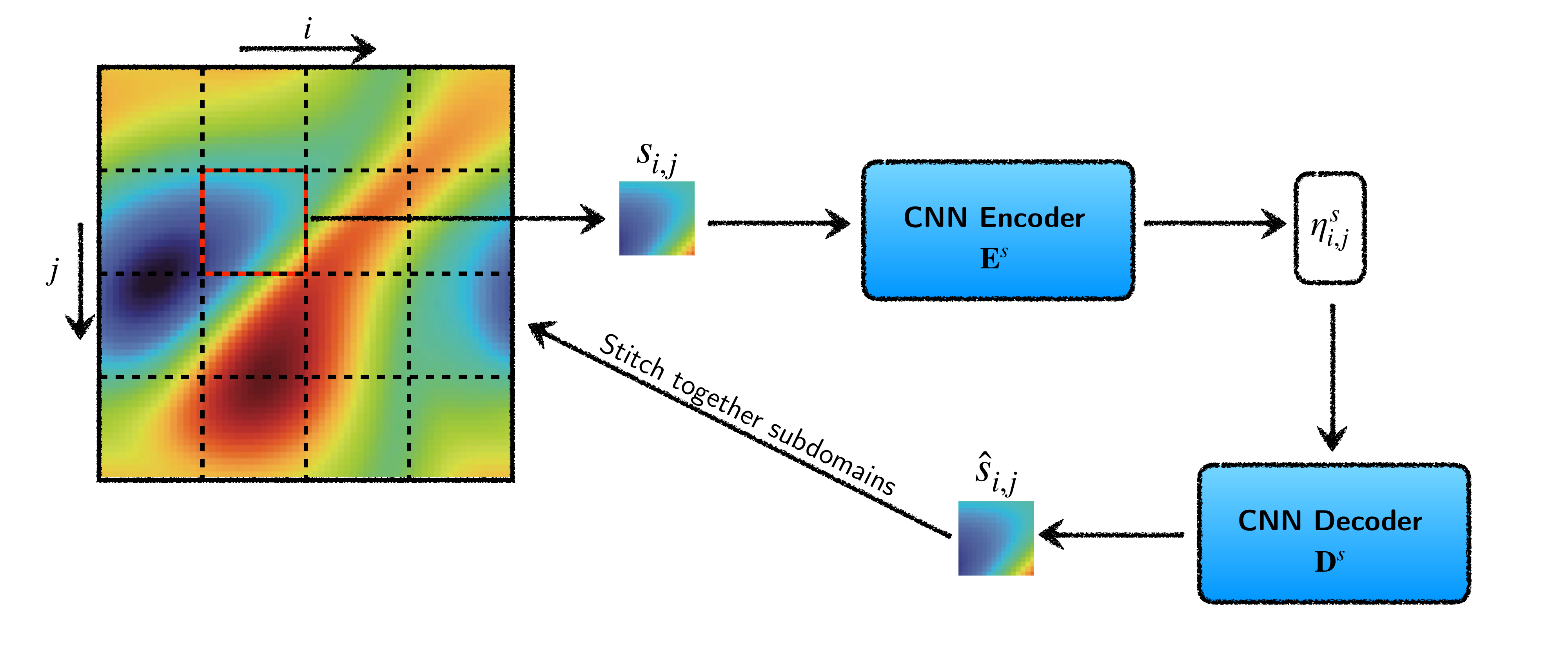}
\caption{Schematic of the structured CNN-based autoencoder used to encode the physical fields to the latent space $\eta$ and to decode back from the latent space $\eta$ to the physical space. Encoding and decoding of condition variables are performed in a similar fashion. For 3-D datasets, a 3-D CNN is used.}
\label{fig:schematic_ae}
\end{figure}

Our methodology decomposes the global computational domain $\Omega_G$ into uniform stuctured subdomains $\Omega_1, \Omega_2, \cdots, \Omega_N$ such that $\Omega_1 \cup \Omega_2 \cdots \cup \Omega_N = \Omega_G$. Each subdomain represents a constant physical size and denotes the local distribution of solution and condition of the PDE distribution, $s_n$ and $c_n$, respectively. A CNN-based autoencoder is used to learn to encode and decode $s_n$ and $c_n$. The trained CNN-based autoencoder determines the compressed latent representations $\eta_n^{s}$ and $\eta_n^{c}$ of the local distribution in each structured subdomain $\Omega_n$. The encoder network $\mathbf{E}$ learns to compress the input distribution in a latent representation as shown in equation \ref{encoder_eq}. The decoder of the network $\mathbf{D}$ learns to reconstruct the original local distribution based on the latent representation as shown in equation \ref{decoder_eq}. The encoder and decoder networks are trained together using a reconstruction objective function with solution/condition fields represented on subdomains. Mean squared error (MSE) is used as the loss function for optimization. During the inference time, encoder and decoder are used seperately to encode $\{s_n, c_n\}$ into $\{\eta^s_n, \eta^c_n\}$ and to decode $\eta^s_n$ obtained from the time-integrator network to get back to physical space $\hat{s}_n$. We also note that the condition and solution autoencoders are trained independently, with each field utilizing its own dedicated autoencoder instance if multiple fields are involved. The schematice of the CNN-based autoencoder operating on subdomains is depicted in figure \ref{fig:schematic_ae}.

\subsection{Robust transient learning with curriculum learning} \label{sec:transient_network}

\begin{figure}[htbp]
\centering
\includegraphics[trim={0.0cm 8cm 0.0cm, 0cm},clip=true,width=1\linewidth]{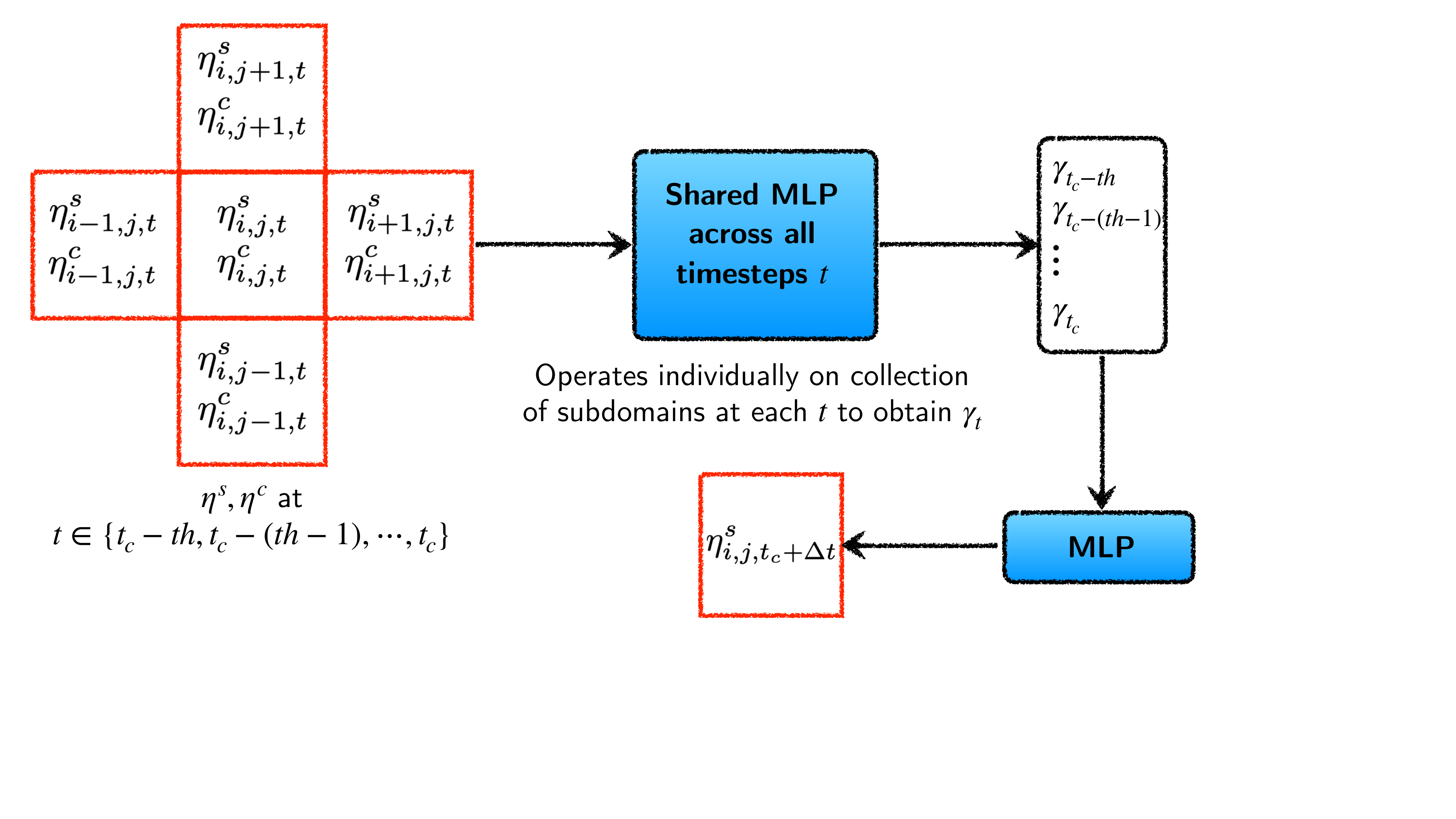}
\caption{Schematic of the time integrator operating in the latent space $\eta$.}
\label{fig:schematic_ti}
\end{figure}

% \begin{itemize}
%     \item Algorithm for training
%     \item Algorithm for inference
% \end{itemize}

{\bf TI} focuses on learning the change of distributions of solution latent embeddings in a subdomain from the given time history of solution and condition latent embeddings in its surroundings. The network follows a local learning principle, i.e., it takes a fixed length $\mathcal{T}_s = th$ and $\mathcal{T}_c = th$ of the history of the local distribution of solution and condition in the latent space, $\{\eta^s_{i, j, \mathcal{T}_s}, \eta^c_{i, j, \mathcal{T}_c}\}$ along with collection of its neighboring subdomains $\mathcal{I}$, as the input. Figure \ref{fig:schematic_ti} presents a 2-D physical domain for the illustration purpose. Here $i, j$ refer to the index of the subdomain in the $x$ and $y$ direction respectively. A subdomain with index $i,j$ has four neighbors at ($i-1,j$), ($i+1,j$), ($i, j-1$), and $(i, j+1)$.

The {\bf TI} network consists of two deep neural networks, each composed of multi-layer perceptons (MLP) as shown in figure \ref{fig:schematic_ti}. We utilize the first network to fuse the local distribution of neighboring subdomains of solution and condition to generate vecotors $\gamma_t$ for each timestep in the time history $t \in \{t_c - th\Delta t,t_c - (th - 1)\Delta t, ...,t_c\}$, i.e., $\gamma_t = F_{spatial}(\eta^s_{\mathcal{I}, t}, \eta^c_{\mathcal{I}, t}; \theta_{F_{spatial}})$. The second network is used to predict the latent embedding of the solution, $\eta_{i,j, t_c+\Delta t}^s$, for the next time step $t_c + \Delta t$ from the generated $\gamma_{\mathcal{T}_c}$ for all previous timesteps, i.e., ${\eta}^s_{i, j, t+\Delta t} = F_{\text{temporal}}(\gamma_{\mathcal{T}_c}; \theta_{F_{\text{temporal}}})$. We again emphasize that for each timestep, all the subdomains in the computational domain are updated before moving to the next timestep.

For training the time integrator $\mathbf{TI}$, we utilize a multi-step objective function to further increase the robustnes of the training process. Rather than updating weights aggresively based on prediction error of one time step, we let {\bf TI } accomplish forward pass on multiple steps and then compute the error of these steps together for weight updates. For our runs, we find that a weight update after accumulation of loss for 10 timesteps provides a good compromise between the robustness of the model and GPU memory requirement for the training. Furthermore, we utilize a curriculum learning (CL) based approach for robust weight updates. The CL training mechanics was originally proposed by Bengio et al. \cite{bengio2015scheduled} in the context of natural language procesisng to mitigate the error accumulation in autoregressive models. Results from Ghule et al. \cite{ghulenlp} showed preliminary succces in using the CL approach for transient PDE modeling, with improvement in generalization and exptrapolation by adapting CL to FNOs. In short, the CL training mechanic randomly picks either the latent embeddings encoded from true solution distribution, $\eta^{s, true}_{\mathcal{I}}$, or the latent embeddings based on the  {\bf TI} prediction, ${\eta}^s_{\mathcal{I}}$, as the input for any timestep in the time history. The probability with which the model's prediction is selected for a given timestep is gradually ramped up as the model training progresses. Initially, the model uses ground truth data as input for stable training for certain number of epochs. As training advances, the model progressively starts using its own predictions as input. 

\subsection{Details of inference methdology} \label{sec:training_inference}

For the inference of each transient PDE sample, we start with latent embeddings of initial history solutions and condtions, \{$\eta^s_t, \eta^c_t$ \}, for $th$ history time steps, $t \in \{0, \Delta t, ..., th \Delta t \}$, encoded by the trained solution and condition encoders, ${\bf E}^{s}$ and ${\bf E}^{c}$, respectively. We apply the forward pass on each subdomain at time step $t_c$ so that the solution latent embeddings $\eta^s_{t_c+\Delta t}$ of sudomains in the entire computational domain ($\Omega_G$) are predicted. With the latest predicted $\eta^s_{t_c+\Delta t}$ added to the time history solutions, we continue unrolling in time to predict the solutions for the next timesteps. The unrolling of solutions over time occurs entirely in the latent space, providing users with the flexibility to decode only the timesteps of interest.

{\bf AE} and {\bf TI} are trained separately following our descriptions in section \ref{sec:mesh_autoencoder} and \ref{sec:transient_network}, respectively. As {\bf TI} learns in the latent space of PDE solutions, we first train the {\bf AE} and use it for encoding and decoding solution fields while training the {\bf TI}. The inference methdology is explained in detail in algorithm \ref{alg:inference_alg}. The methodology starts with the intial condition and solution distributions of the first $th$ timesteps. The transient-CoMLSim discretizes the entire physical domain $\Omega_{G}$ into subdomains $\Omega_{1}, \Omega_{2}, ..., \Omega_{N}$.  The trained solution encoder ${\bf {E^s}}$ and condition encoder ${\bf {E^c}}$ are used to encode the solution and condition distribution of each subdomain $\Omega_{n}, n \in \{1, 2, ..., N \}$ into $\eta^s_{n, t}$ and $\eta^c_{n, t}$, for the first $th$ time steps $t \in \{0, 1, ..., th\}$. After obtaining all the latent embeddings of the initial history of time steps, the $\mathbf{TI}$ can predict the solution distribution of the next steps in an autoregressive way until the last time step. For each time step, the trained encoder produces the condition embedding $\eta^c$ from the input data. The trained {\bf TI} predicts the solution distribution of each subdomains for the next time step $t_c+ \Delta t$ following equation \ref{time_int_eq}. In equation \ref{time_int_eq}, the spatial information needed to update the central subdomain for the next timestep is obtained from neighboring subdomains, as illustrated in figure \ref{fig:schematic_ti}. At each timestep, we predict the solution encoding across all subdomains in the computational domain. This method facilitates the propagation of information throughout the entire domain. After obtaining the encodings of the entire time sequence, the decoder $\mathbf{D^s}$ decodes the latent embeddings $\eta^s_{n, th}$ to  $\eta^s_{n, th+T \Delta t}$ into solution distribution patches and the subdomains are put back into the original computational domain to compose the solution predictions at timesteps $th, th+\Delta t, \cdots, th + T \Delta t$. At inference, one can also impose periodic and Dirichlet boundary condition in the latent space, respectively, as follows: 

\begin{equation}
    \eta^s_{\mathrm{start}, t_c} = \eta^s_{\mathrm{end}, t_c},
    \label{eq:periodic_bc}
\end{equation}

\begin{equation}
    \eta^s_{\mathrm{start}, t_c} = \eta^s_{\mathrm{start}, t_c} = \eta^s_{\mathrm{0}}.
    \label{eq:dirichlet_bc}
\end{equation}

In the equations above, $\mathrm{start}$ and $\mathrm{end}$ denote the first and last subdomains along the axis where the boundary condition is applied.

\begin{algorithm}[H]
\caption{Inference methodology of transient-CoMLSim.} \label{alg:inference_alg}
\begin{algorithmic}[0]
\State Domain Decomposition: Domain $\Omega_{G}$ into $\Omega_{1}, \Omega_{2}, \cdots, \Omega_{N}$
\For{time step $t \in \{0, 1, ..., th\}$} \Comment{Initialize with the condition and solution history} 
    \For{all subdomain $n \in \{1, 2, ..., N\}$}
        \State Encode initial conditions: $\eta^c_{n, t}= E^c(c_{n,t})$ 
        \State Encode initial solutions: $\eta^s_{n, t} = E^s(s_{n,t})$
    \EndFor
\EndFor
\State $t_c \gets th$
\While{time step $t_c < \text{final time step} \ th + T \Delta t$}
    \For{subdomains $\Omega_i \in \Omega_G$}
        \State Fetch history of latent vectors in neighborhood of $\Omega_n$: $\eta^s_{\mathcal{I}, \mathcal{T}_s}$ , $\eta^c_{\mathcal{I}, \mathcal{T}_c}$ 
        \State Compute solution encodings of $\Omega_n$ at $t_c + \Delta t$ by {\bf TI}: $\eta^s_{n, t_c+\Delta t} = \mathbf{TI}(\eta^s_{\mathcal{I}, \mathcal{T}_s}, \eta^c_{\mathcal{I}, \mathcal{T}_c}; \theta)$ 
    \EndFor
    \State $t_c \gets t_c + \Delta t$ 
\EndWhile
\State Decode PDE solutions after inference in the latent space for $th$ to $th + T \Delta t$.  
\end{algorithmic}
\end{algorithm}

% \begin{algorithm}
% \caption{An algorithm with caption}\label{alg:cap}
% \begin{algorithmic}
% \Require $n \geq 0$
% \Ensure $y = x^n$
% \State $y \gets 1$
% \State $X \gets x$
% \State $N \gets n$
% \While{$N \neq 0$}
% \If{$N$ is even}
%     \State $X \gets X \times X$
%     \State $N \gets \frac{N}{2}$  \Comment{This is a comment}
% \Else
%     \State $y \gets y \times X$
%     \State $N \gets N - 1$
% \EndIf
% \EndWhile
% \end{algorithmic}
% \end{algorithm}

\subsection{Description of datasets}

\begin{table}[ht]
\centering
\caption{Summary of datasets employed in this study. $N_d, N_\textrm{var}$ denote the number of dimensions and number of variables, respecitvely. $N_x, N_y, N_z, N_t$ correspond to the number of grid points in $x, y, z, t$ directions, respectively. $N_{\textrm{train}}$ and $N_\textrm{test}$ denote the number of training and test samples, respectively.}
\label{tab:dataset_description}
\begin{tabular}{@{}ccccccccc@{}}
\toprule
Dataset  & $N_d$ & $N_\textrm{var}$ & $N_x$ & $N_y$ & $N_z$ & $N_t$ & $N_{\textrm{train}}$ & $N_{\textrm{test}}$  \\
\midrule
2-D Shallow Water  & 2 & 1 & 128 & 128 & -- & 100 & 800 & 100 \\
2-D Diffusion-Reaction  & 2 & 2 & 128 & 128 & -- & 100 & 800 & 100 \\
2-D vortex-flow & 2 & 1 & 64 & 64 & -- & 100 & 800 & 100 \\
3D Additive Manufacturing   & 3 & 2 & 16 & 200 & 200 & 400--1000 & 48 & 6 \\
%2-D Cylinder Flow  & 2 & 3 & 256 & 64 & -- & 120 & & \\
\bottomrule
\end{tabular}
\end{table}

We test our transient-CoMLSim model on four different datasets, each of which are governed by a different set of PDEs, as shown in table \ref{tab:dataset_description}. A brief description of each dataset is provided below: 

\textbf{2-D Shallow-Water Equation}: 
The shallow water equation (SWE) dataset is sourced from PDEBench \citep{takamoto2022pdebench} and offers a framework for simulating free surface flows. The dataset represents a 2-D radial dam break scenario, initializing the circular bump's initial height at the center of the domain as:
\begin{equation}
    h(t=0, x, y) = 1.0 \ \forall \ r \geq r_c \  \textrm{and} \   h(t=0, x, y) =2.0 \  \forall \  r < r_c,
\end{equation}
where $r_c$ is uniformly sampled from the uniform distribution $\mathcal{U}(0.3, 0.7)$. These simulations accurately portray a radially propagating shock front as time $t$ progresses. The dataset consists of a single variable $h$, denoting the height of the water level in the 2-D domain at different timesteps $t$. The 2-D SWE PDE is used to model complex phenomena like tsunamis and flooding events, where mass and momentum conservation must be maintained, even across shocks. These characteristics make 2-D SWE dataset particularly difficult, as its requires accurate handling of nonlinear dynamics and discontinuities in the temporal 2-D solution.

\textbf{2-D Diffusion-Reaction Equation}: The 2-D diffusion-reaction equation, also taken from PDEBench \cite{takamoto2022pdebench}, consists of two non-linearly coupled variables, $u(x,y,t)$ and $v(x,y,t)$, that evolve according to the following parabolic PDE equations, 

\begin{equation}
    \frac{\partial u}{\partial t} = D_u \Big(\frac{\partial^2 u}{\partial x^2} + \frac{\partial^2 u}{\partial y^2} \Big) + R_u, \frac{\partial v}{\partial t} = D_v \Big(\frac{\partial^2 v}{\partial x^2} + \frac{\partial^2 v}{\partial y^2} \Big) + R_v.
\end{equation}

Here, $R_u = u - u^3 - k - v$ and $R_v  = u - v$ with $k = 5 \times 10^{-3}$, as given in Takamoto et al. \cite{takamoto2022pdebench}. The intial conditions for both $u$ and $v$ are generated as Gaussian noise sampled from $\mathcal{N}(0,1)$. Thus, the dataset consists of two variables that evolve together parabolically in space and time. The 2-D diffusion-reaction problem is particularly challenging because it demands that a machine learning model accurately capture the nonlinear, time-dependent coupling between two interacting variables.

\textbf{2-D vortex-flow}: The 2-D vortex-flow dataset represents spatiotemporally evolving 2-D Navier Stokes equations on a period domain:

\begin{equation}
    \frac{\partial \omega}{\partial t} + u \frac{\partial \omega}{\partial x} + v \frac{\partial \omega}{\partial y} = \frac{1}{Re}\Big(\frac{\partial^2 \omega}{\partial x^2} + \frac{\partial^2 \omega}{\partial y^2} \Big) + f(x,y),
\end{equation}

\begin{equation} 
    \frac{\partial u}{\partial x} + \frac{\partial v}{\partial y} = 0,
\end{equation}

where $\omega$ is the vorticity and $u,v$ are the velocity components. In our work, we set $f = 0.1(\mathrm{sin}(2\pi(x+y)) + \mathrm{cos}(2\pi(x+y))$ and $Re = 1000$. The vorticity field $\omega_0 (x,y)$ is initialized at $t=0$ using Gaussian random field for the entire dataset. In our dataset for the 2-D vortex-flow, we generate data at $256 \times 256$ resolution on a $[0,1] \times [0,1]$ square box and downsample it to $64 \times 64$ grid. We are interested in obtaining the spatiotemporal evolution of $\omega (x,y,t)$.  The 2-D vortex flow is a canonical dataset used widely in testing physics-based ML models \cite{li2020fourier,kochkov2021machine,kovachki2023neural,bar2019learning}.

\textbf{3-D Additive Manufacturing}: The 3D additive manufacturing dataset illustrates an industrial scenario where the temperature ($T$) distribution of a metal plate undergoes changes as a laser source moves over it during the 3-D printing process. This evolving temperature field, influenced by laser motion, can induce rapid heating and cooling, resulting in the generation of strain and stress within a manufactured part. Consequently, these factors can significantly impact the structural integrity of the final product. The PDE equation that governs the temperature variation is:

\begin{equation}
    {\rho C}\frac{\partial T}{\partial t} = k \Big(\frac{\partial^2 T}{\partial x^2} + \frac{\partial^2 T}{\partial y^2} + \frac{\partial^2 T}{\partial z^2}\Big) + Q(x,y,z,t),
\end{equation}

where $Q(x,y,z,t)$ represents the moving laser source and $\alpha = k/\rho C$ is the thermal diffusivity of the metal. Here, $k, \rho$ and $C$ are the thermal conductivity, density, and specific heat, of the material. In our work, $k, \rho, C,$ and laser power parameters stay constant while the laser motion pattern changes across the different samples. Hence, $N_t$ varies across different samples. For training, we keep $N_t = 400$ across all training samples. However, at inference, we let the model run for the entire duration of the actual simulation. This can lead to $N_t$ going as large at $1500-1600$ at inference -- approximately $4\times$ the temporal window seen by the model during training. It is important to note that, given the three-dimensional nature of the additive manufacturing dataset, it is prohibitively expensive to generate $N_{train}$ or $N_{test}$ of $\sim O(100)$. Hence, during training, we augment 12 simulations to generate 48 training samples by performing two reflections along the $x-z$ and $y-z$ planes and one $90^\circ$ rotation about the $z$-axis. During inference, we test the model's performance on 6 non-augmented simulations with varying time histories and spatial domain sizes, as mentioned in table \ref{tab:additive_details}.

\begin{table}[ht]
\centering
\caption{Summary of simulation parameters for test samples of 3-D additive manufacturing dataset.}
\label{tab:dataset_description}
\begin{tabular}{@{}ccccccccccccccccccccccc@{}}
\toprule
Index & & & & $N_x$ & & & & $N_y$ & & & & $N_z$ & & & & $N_t$  \\
\midrule
AM1 & & & & 200 & & & & 200 & & & & 16 & & & & 298 \\
AM2 & & & & 200 & & & & 200 & & & & 16 & & & & 668 \\
AM3 & & & & 200 & & & & 200 & & & & 16 & & & & 417 \\
AM4 & & & & 296 & & & & 296 & & & & 16 & & & & 1600 \\
AM5 & & & & 544 & & & & 544 & & & & 16 & & & & 601 \\
AM6 & & & & 104 & & & & 1320 & & & & 16 & & & & 601 \\

\bottomrule
\end{tabular}
\label{tab:additive_details}
\end{table}

\subsection{Description of baseline models and error metric}
Our transient-CoMLSim is compared against the following baseline models to assess its performance. 

\textbf{U-Net}: The U-Net model \citep{ronneberger2015u} along with pushforward technique \citep{brandstetter2022message} is used, similar to the implementation of Takamoto et al. \cite{takamoto2022pdebench}. As discussed in Takamoto et al. \cite{takamoto2022pdebench}, the autoregressive method results into training instability for U-Net. To avoid this, we have used the pushforward technique to train the model, following Takamoto et al.\cite{takamoto2022pdebench}. In our study, we use a pushforward of 20 timesteps. The gradients are calculated only for the last 20 timesteps. At training, the model rolls out predictions similar to an autoregressive model, but only the last 20 steps are used for backpropagation. The U-Net model has 7765057 trainable parameters. The U-Net model takes a time history of $th=10$ steps as input to predict the next timestep. 

\textbf{Fourier Neural Operator}: FNO \citep{li2020fourier} is an established physics-based ML approach that learns the forward propagator for PDEs. The implementation of FNO in this study is same as in Takamoto et al. \cite{takamoto2022pdebench} with initial $10$ timesteps used as input to the model while the model unrolls for the remaining $N_t - 10$ timesteps autoregressively. The model has 465557 parameters that can be trained.

We use the normalized root mean squared error (nRMSE) to quantify the accuracy of predictions for different PDEs across various models. 

\begin{equation}
    \mathrm{nRMSE} = \frac{1}{N_{test} \cdot (N_t - th) \cdot N_{var}} \sum^{N_{test}}_{n=1} \sum^{N_{var}}_{nvar=1}\sum^{N_t}_{t=th+1} \frac{\norm{\mathbf{u}_{pred,t,nvar,n} - \mathbf{u}_{gt,t,nvar,n}}_2}{\norm{\mathbf{u}_{gt,t,nvar,n}}_2}
\label{eq:nrmse}
\end{equation}

nRMSE normalizes the RMSE based on the $\ell^2$ norm of the ground truth data, making it dimensionless and comparable across different datasets. This  ensures that the error metric remains consistent, irrespective of the scale of the data, providing a more meaningful assessment of model performance.

\section{Results}\label{sec:results}
 
This section presents results obtained from our experiments, focusing on the evaluation of the accuracy, generalizability, and extrapolation capability of our framework. We conduct comparisons with the benchmark models outlined in the preceding section.

\subsection{Accuracy comparison with baselines for the interpolation task}\label{sec:interpolation}

\begin{table}[ht]
\caption{Comparison of nRMSE among transient-CoMLSim, FNO and U-Net models on the four datasets. For FNO and transient-CoMLSim, gradient updates are performed after every 10 timesteps. FNO and transient-CoMLSim both are trained with 10 timesteps as history and U-Net is trained with a pushforward timestep window of 20.}\label{tab:tab1}
\centering
\begin{tabular}{c c c c c c}
\toprule
Dataset & Transient-CoMLSim & FNO & U-Net \\
\midrule
2-D Shallow Water & $1.96 \times 10^{-3}$ & $3.10 \times 10^{-3}$ & $8.54 \times 10^{-2}$ \\
2-D Diffusion-Reaction & $5.32 \times 10^{-2}$ & $2.07 \times 10^{-1}$ & $3.73 \times  10^{-1}$ \\
2-D Vortex Flow & $9.67 \times 10^{-2}$ & $2.09 \times 10^{-1} $ & $6.31 \times 10^{-1}$ \\
3-D Additive Manufacturing  & $5.47 \times 10^{-2}$ & $-$ & $-$   \\
\bottomrule
\end{tabular}
\label{tab:accuracy_over_baselines}
\end{table}

\begin{figure}[htbp]
\centering
\includegraphics[width=1.2\linewidth]{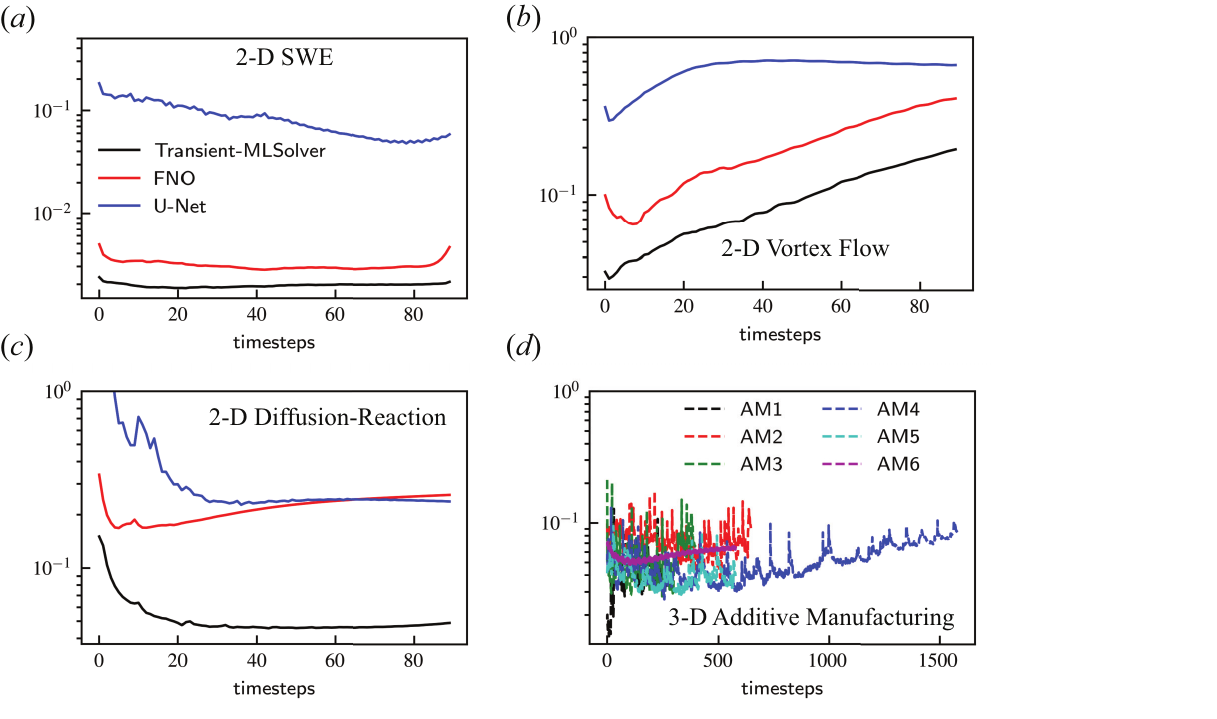}
\caption{Evolution of nRMSE as a function of timesteps for the four datasets. For panels $(a)-(c)$, all models are trained until $T=100$ timesteps and are inferred from $t=11$ to $100$ on the test samples. In panel $(d)$, models are trained for 400 timesteps and inferred until the end of the time horizon for the respective simulation. As mentioned in the text, we were unable to fit FNO or U-Net on the additive manufacturing dataset using our compute resources. Therefore, panel $(d)$ only shows the nRMSE evolution for the transient-CoMLSim.}
\label{fig:nrmse_interp}
\end{figure}

Table \ref{tab:accuracy_over_baselines} presents the nRMSE averaged over all inferred timesteps and data samples for the four PDEs described in the previous section. For all the PDEs, we observe that transient-CoMLSim performs significantly better than both the FNO and U-Net baselines. Each dataset presents unique challenges: (a) the 2-D SWE dataset requires capturing a propagating wave front, (b) the 2-D diffusion-reaction dataset involves complex non-linear coupling between the $u$ and $v$ variables, (c) the 2-D vortex-flow dataset is chaotic in nature, and (d) the 3-D additive manufacturing dataset involves capturing the complex laser motion patterns for extended time durations. Transient-CoMLSim outperforms the benchmark models in all these cases. An important point to note is that both transient-CoMLSim and FNO perform optimization steps after accumulating gradients for 10 timesteps. While FNO outperformed transient-CoMLSim on the 2-D vortex-flow dataset when FNO's weight optimization was performed after accumulating gradients for up to 100 timesteps, this approach might be restrictive on (a) architectures with limited GPU memory and (b) for PDEs that are three-dimensional in nature. In such cases, transient-CoMLSim remains a more attractive option, even though FNO could potentially achieve better performance with larger gradient accumulation. For other datasets, FNO underperformed compared to transient-CoMLSim even when FNO weight optimization was performed after gradient accumulation for 100 timesteps (not shown here). The subdomain size for all the runs of transient-CoMLSim is set as $8\times8$ for 2-D datasets and $8\times8\times8$ for 3-D dataset. The size of the latent embedding is set to $16$ for 2-D datasets and $8$ for the 3-D additive manufacturing dataset. In the limit of larger subdomain sizes, our framework begins to resemble FNO or U-Net-like architectures, capturing broader spatial features but losing some of the granularity that smaller subdomains provide. Conversely, as the subdomain size decreases, the model increasingly behaves like a traditional solver, effectively preserving local details but at the cost of reduced data compression. Through our experiments, we found that a subdomain size of 8 in each direction offers an optimal balance, compressing the dataset efficiently while maintaining accuracy. The subdomain size of $8$ in each direction effectively captured the essential dynamics of the system without excessively smoothing the small-scale features

Due to the significantly larger domain size and the three-dimensional nature of the additive manufacturing dataset compared to the 2-D datasets, fitting the FNO and U-Net model and data on the available GPU was challenging. Furthermore, the laser power source in the 3-D additive manufacturing dataset is highly localized for any given timestep, making it unsuitable for spatial subsampling to fit on available compute resources for the FNO and U-Net models. The methodology used in PDEBench \cite{takamoto2022pdebench} calculates the loss for each time step and accumulates it to compute gradients after the entire trajectory is rolled out. Given the large domain size and number of roll-out steps, we encountered out-of-memory (OOM) errors for both FNO and U-Net, even with a batch size of one. Hence, with the available resources, we could not train the baselines as per the above methodology for the 3-D additive manufacturing dataset. To overcome the OOM error, we modified the training mechanism by calculating gradients for each time step instead of all at once. This adjustment allowed us to train the models with a batch size of two samples. The per epoch time for this setup is approximately 50 minutes, making training for 500 epochs, as suggested in FNO \citep{li2020fourier}, impractical. In contrast, the method proposed in our paper is well-suited to handle such industrial use cases. Hence, we could not report the baseline comparison for the 3-D additive manufacturing dataset in table \ref{tab:accuracy_over_baselines}.

Figure \ref{fig:nrmse_interp}(a,b,c) show the evolution of nRMSE over time for 2-D datasets when all the models are trained until $T = 100$ timesteps. For this plot, we average $\mathrm{nRMSE}$ over all the test data samples for a given time $t$. Transient-CoMLSim and FNO both are trained with time history of 10 timesteps. Between the U-Net with pushforward and FNO, FNO performs better and accumulates less error. However, transient-CoMLSim significantly outperforms both on the test dataset, demonstrating that it is a better temporal interpolator framework compared to the baselines. 

Since the 3-D additive manufacturing dataset includes test samples that evaluate the model's extrapolation capability to larger domain sizes and longer timesteps, we present the nRMSE evolution for all test cases in figure \ref{fig:nrmse_interp}(d). It can be observed that our model consistently predicts within a $5-10\%$ error range for all domain sizes and timesteps, even though it was trained on only 400 timesteps.

\begin{figure}[htbp]
\centering
\includegraphics[width=1\linewidth]{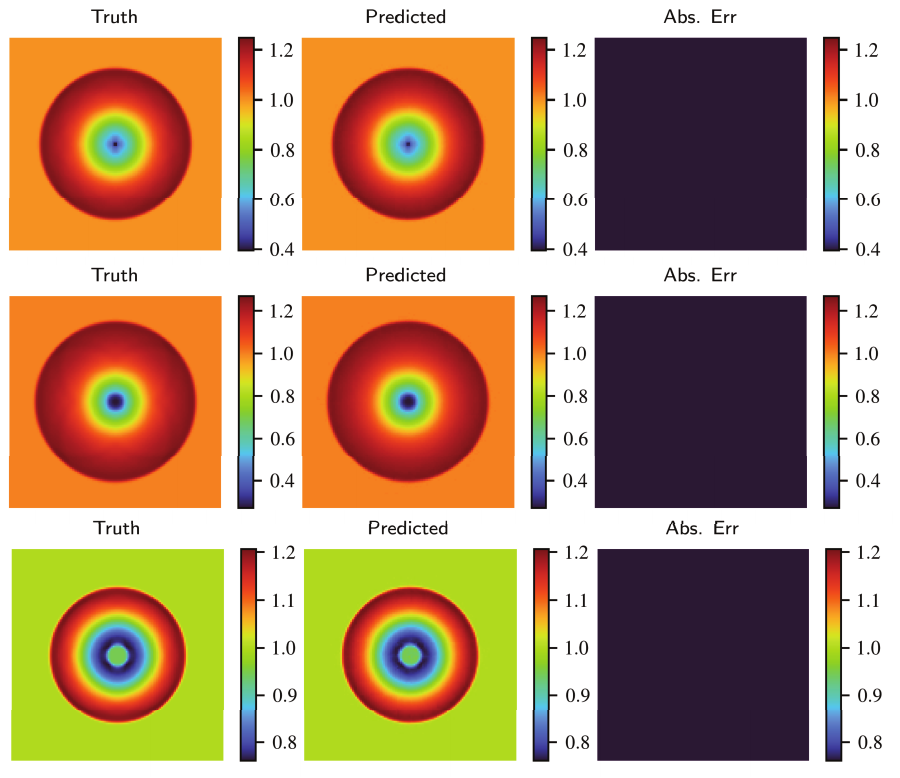}
\caption{Randomly selected three test samples at $t=100$ randomly selected from 2-D shallow water dataset.}
\label{fig:swe_interp_contours}
\end{figure}

\begin{figure}[htbp]
\centering
\includegraphics[width=1\linewidth]{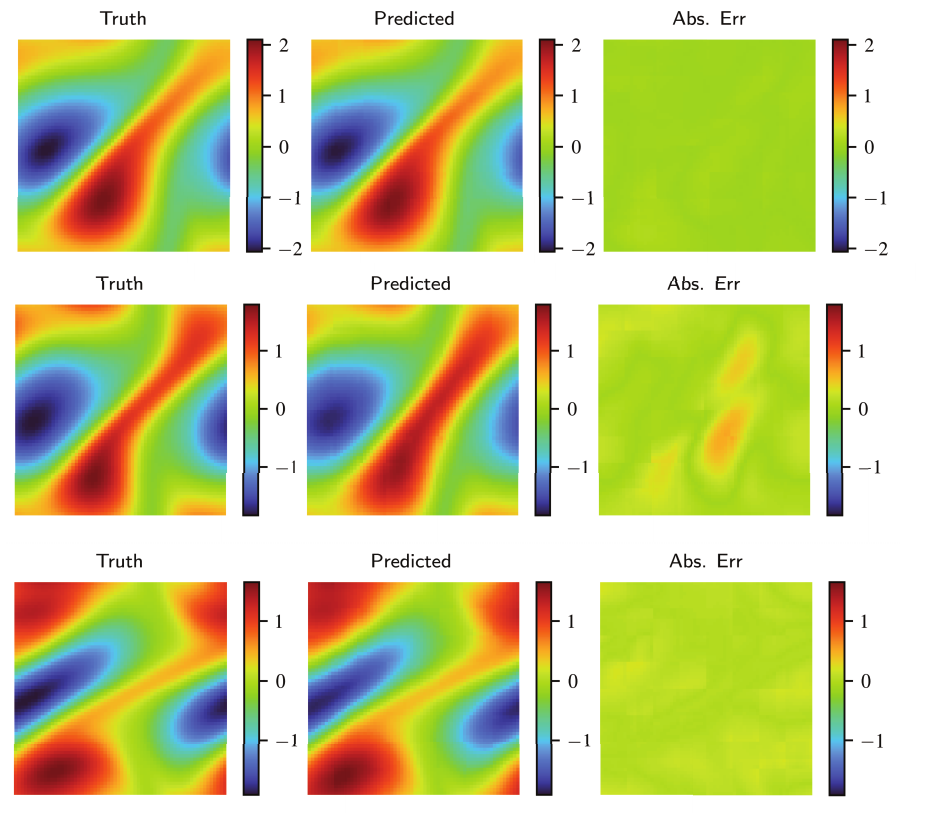}
\caption{Randomly selected three test samples at $t=100$ from 2-D vortex-flow dataset.}
\label{fig:fno_interp_contours}
\end{figure}

\begin{figure}[htbp]
\centering
\includegraphics[width=1\linewidth]{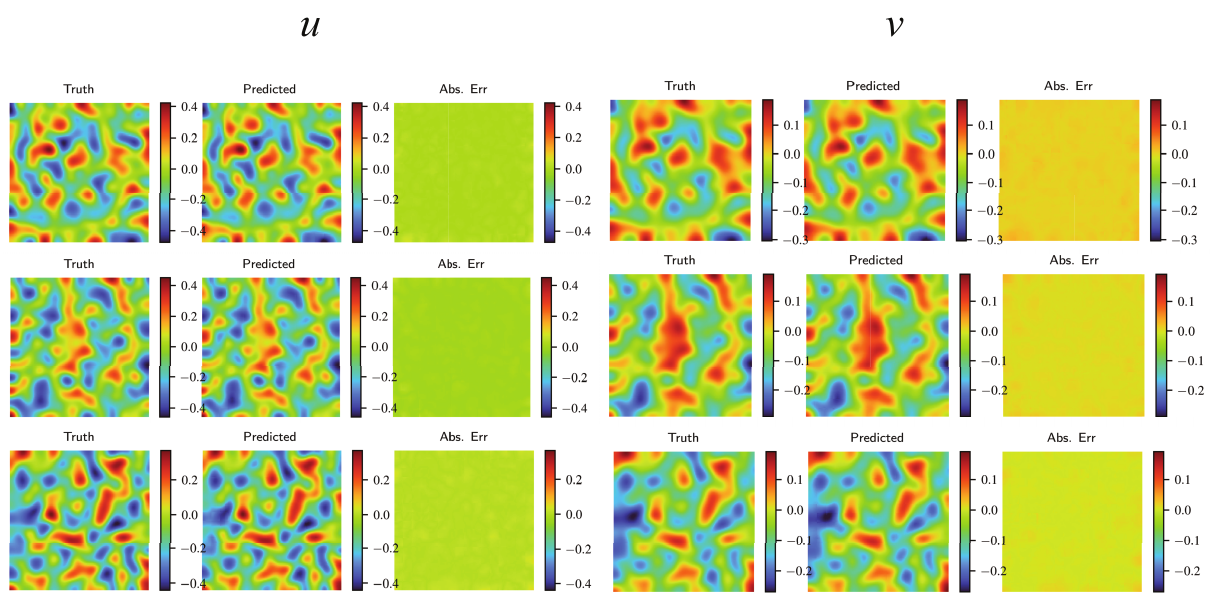}
\caption{Randomly selected three test samples at $t=100$ from 2-D diffusion-reaction dataset.}
\label{fig:diff-reac_interp_contours}
\end{figure}

Figures \ref{fig:swe_interp_contours}, \ref{fig:fno_interp_contours}, \ref{fig:diff-reac_interp_contours} present three randomly selected contours obtained using transient-CoMLSim at last timesteps from the 2-D shallow water, 2-D vortex-flow, and 2-D diffusion-reaction, respectively. From the contours, it is evident that the transient-CoMLSim captures the physics of the respective 2-D datasets quite accurately throughout the course of their respective temporal evolution. Appendix \ref{sec:appendix} presents animations of three randomly selected test samples, showing the time evolution of solution variables for the three 2-D PDE datasets (figures \ref{fig:output_swe1}--\ref{fig:output_diff3}).

\subsection{Extrapolation to unseen timesteps and bigger domain sizes} \label{sec:extrapolation}

\begin{table}[ht]
\caption{Performance of the different models on unseen timesteps. Models are trained for $t \in [10, 70]$ and are inferred on $t \in [10, 100]$.}\label{tab:tab2}
\centering
\begin{tabular}{c c c c c c c}
\toprule
Dataset & Transient-CoMLSim & FNO \\
\midrule
2-D Shallow Water & $6.42 \times 10^{-3}$ & $1.40 \times 10^{-2}$  \\ 
2-D Diffusion-reaction & $3.97 \times 10^{-2}$ &  $5.65 \times 10^{-2}$  \\
2-D vortex-flow & $8.37 \times 10^{-2}$ & $1.31 \times 10^{-1}$  \\
\bottomrule
\end{tabular}
\label{tab:extrapolation}
\end{table}

To test the performance on unseen timesteps, we train the transient-CoMLSim model on timesteps $t \in [10, 70]$ and unroll the model up to $T = 100$ at inference. Table \ref{tab:extrapolation} presents the nRMSE, calculated from \ref{eq:nrmse}, for SWE, diffusion-reaction, and vortex-flow datasets. Transient-CoMLSim is compared against FNO. We do not perform the extrapolation runs for the U-Net because, from table \ref{tab:accuracy_over_baselines}, we infer that FNO outperforms U-Net in terms of accuracy. Similar to table \ref{tab:accuracy_over_baselines}, time history and number of timesteps over which gradient accumulation is done are set to 10 for both models. Table \ref{tab:extrapolation} shows that the transient-CoMLSim outperforms FNO in terms of its ability to extrapolate to unseen timesteps during inference.

\begin{table}[ht]
\caption{Performance of the transient-MLSolver on 3-D additive manufacturing dataset. Model performs temporal extrapolation for samples AM2 and AM3 and both spatial and temporal extrapolation for AM$4-6$, as discussed in table \ref{tab:additive_details}.}\label{tab:tab2}
\centering
\begin{tabular}{c c c c c c c c c c c}
\toprule
Index & & & & $N_t$  & & & & Transient-CoMLSim \\
\midrule
AM1  & & & & 298  & & & & $4.24 \times 10^{-2}$ \\ 
AM2  & & & & 668  & & & & $7.02 \times 10^{-2}$ \\ 
AM3  & & & & 417  & & & & $5.07 \times 10^{-2}$ \\
AM4  & & & & 1600 & & & & $5.16 \times 10^{-2}$ \\
AM5  & & & & 601  & & & & $3.97 \times 10^{-2}$ \\
AM6  & & & & 601  & & & & $5.80 \times 10^{-2}$ \\
\bottomrule
\end{tabular}
\label{tab:extrapolation_additive}
\end{table}

\begin{figure}[htbp]
\centering
\includegraphics[width=0.9\linewidth]{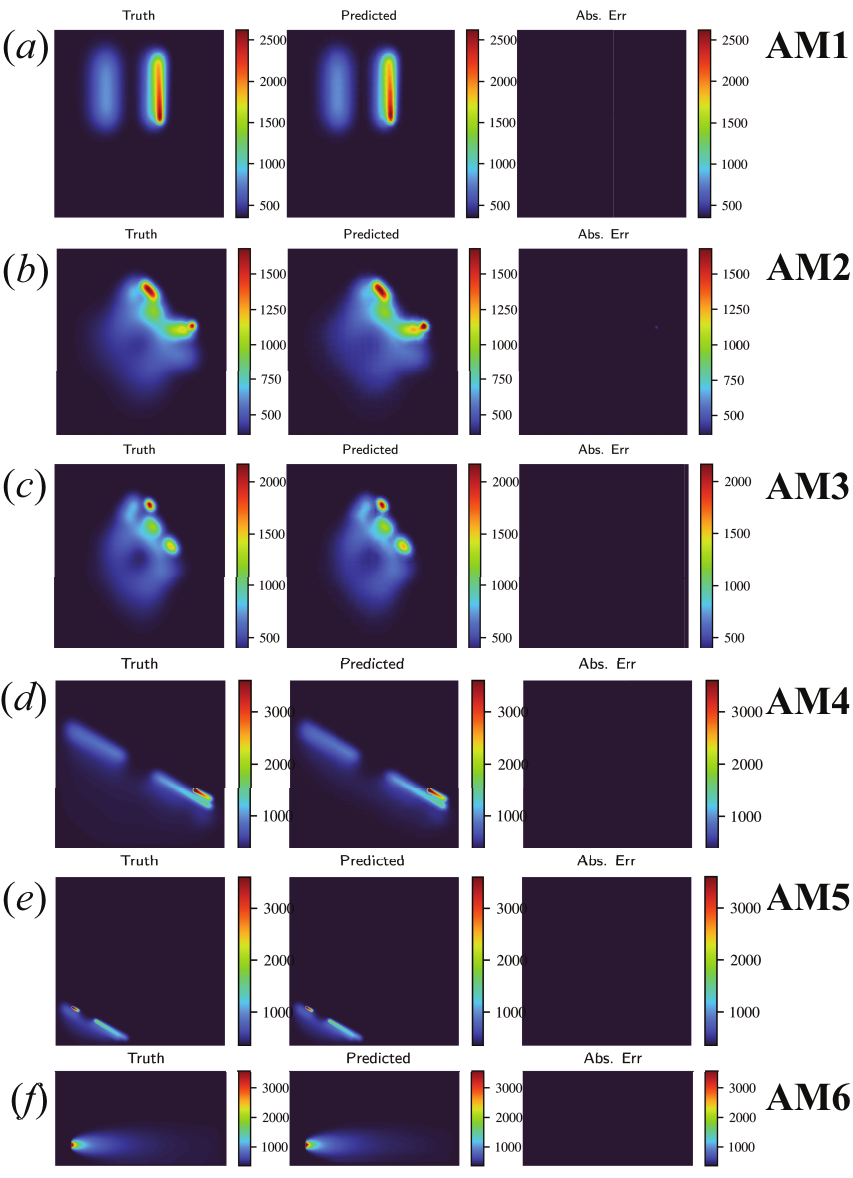}
\caption{Test samples at the end of their respective simulation times for the 3-D additive manufacturing case. Temperature contour at the top-most plane is shown.}
\label{fig:additive_contours}
\end{figure}

\begin{figure}[htbp]
\centering
\includegraphics[width=1\linewidth]{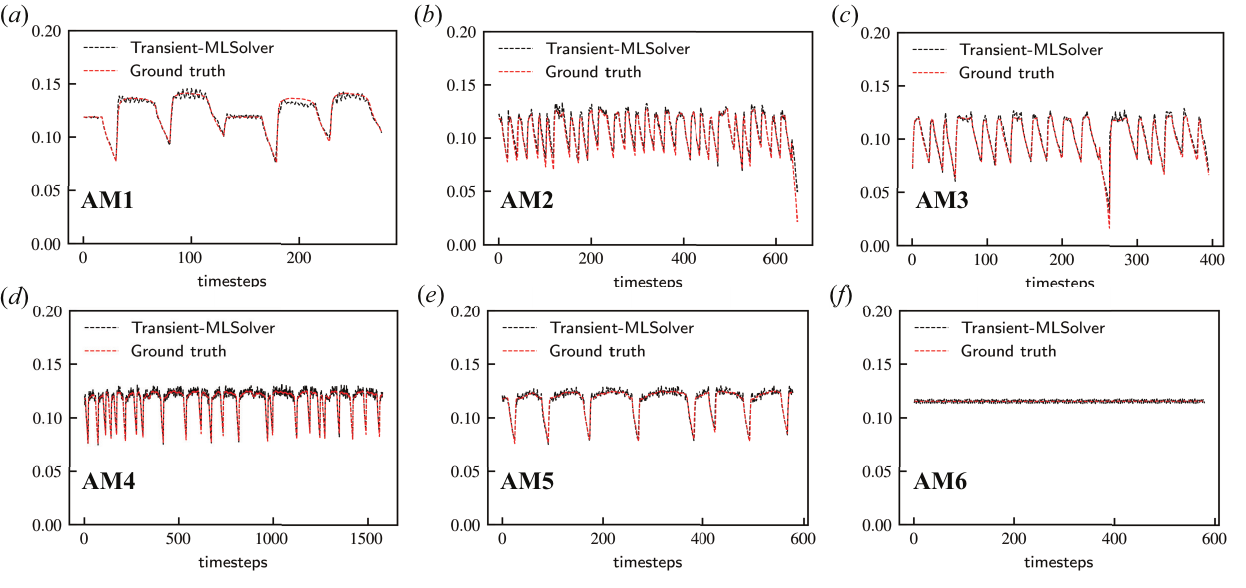}
\caption{Comparison between the ground truth and transient-CoMLSim prediction of melt pool depth as a function of timesteps for test data samples of 3-D additive manufacturing case.}
\label{fig:additive_mp_plots}
\end{figure}

Table \ref{tab:extrapolation_additive} presents the nRMSE of the transient-CoMLSim model on the six test datasets of the 3-D additive manufacturing case. The number of timesteps differ from that used during training -- $N_t = 400$ during training for the 3-D additive manufacturing case. Additionally, datasets AM4, AM5, and AM6 also test the model's ability to extrapolate to larger spatial domains, as discussed in the context of table \ref{tab:additive_details}. Table \ref{tab:extrapolation_additive} demonstrates that the model shows remarkable accuracy for all six test samples, effectively extrapolating to unseen timesteps and domain sizes. Figure \ref{fig:additive_contours} presents the temperature contours at the top-most plane for all six test samples, at the end of their respective simulation time.  The model faithfully captures the spatial distribution of the temperature in the 3-D additive manufacturing case, similar to its 2-D counterparts that are discussed in the previous section. Figure \ref{fig:additive_mp_plots} shows the temporal evolution of the melt pool depth, a global quantity, for all six test samples. In the context of additive manufacturing, the melt pool depth is the depth to which the temperature goes beyond melting point during simulation. Transient-CoMLSim demonstrates remarkable accuracy in capturing the temporal trend of the melt pool depth for all six test cases. Animations of the temperature field evolution on the topmost plane for test cases $\mathrm{AM}1-\mathrm{AM}5$ are shown in figures \ref{fig:am1}--\ref{fig:am5} in appendix \ref{sec:appendix}.

\subsection{Ablation studies for transient-CoMLSim on vortex-flow dataset} \label{sec:ablation_study}

% \begin{table}
% \caption{$th$ ablation on vortex-flow dataset.}\label{tab:tab3}
% \centering
% \begin{tabular}{l c c c c c c}
% \toprule
% Dataset & $th=1$ & $th=5$ & $th=10$ & $th=20$ \\
% \midrule
% 2-D vortex-flow & $7.67 \times 10^{-1}$ & $2.05 \times 10^{-1}$ & $9.67 \times 10^{-2}$ & $3.82 \times 10^{-2}$ \\
% \bottomrule
% \end{tabular}
% \end{table}

% \begin{table}
% \caption{Training window ablation.}\label{tab:tab3}
% \centering
% \begin{tabular}{l c c c c c c}
% \toprule
% Dataset & $t \in [10, 100]$ & $t \in [10, 70]$ & $t \in [10, 40]$ \\
% \midrule
% 2-D vortex-flow & $9.67 \times 10^{-2}$ & $8.37 \times 10^{-2}$ & $1.90 \times 10^{-1}$ \\
% \bottomrule
% \end{tabular}
% \end{table}

% \begin{table}[ht]
% \centering
% \caption{Ablation studies based on $\mathrm{nRMSE}$ loss demonstrating the impact of varying parameters on 2-D vortex-flow: latent size, training time window, and time history.}
% \label{tab:combined_2-D_vortex_flow_ablation}
% \begin{tabular}{@{}lccc@{}}
% \toprule
% Latent size ($l$) & $l = 8$ & $l = 16$ & $l = 24$ \\
% \midrule
% Test Error & $8.57 \times 10^{-2}$ & $9.67 \times 10^{-2}$ & $7.69 \times 10^{-2}$ \\
% \bottomrule
% \toprule
% Time history ($th$) & $th = 5$ & $th = 10$ & $th = 20$ \\
% \midrule
% Test Error & $2.05 \times 10^{-1}$ & $9.67 \times 10^{-2}$ & $3.82 \times 10^{-2}$ \\
% \bottomrule
% \toprule
% Training window & $t \in [10, 100]$ & $t \in [10, 70]$  & $t \in [10, 40]$ \\
% \midrule
% Test Error & $9.67 \times 10^{-2}$ & $8.37 \times 10^{-2}$ &  $1.90 \times 10^{-1}$ \\
% \bottomrule
% \end{tabular}
% \end{table}

\begin{table}[ht]
\centering
\caption{Ablation studies based on $\mathrm{nRMSE}$ loss demonstrating the impact of varying parameters on 2-D SWE: latent size, training time window, and time history.}
\label{tab:combined_2-D_swe_ablation}
\begin{tabular}{@{}cccccccc@{}}
\toprule
Latent size ($l$) & $l = 8$ & $l = 16$ & $l = 24$ \\
\midrule
Test Error & $1.98 \times 10^{-3}$ & $1.96 \times 10^{-3}$ & $1.95 \times 10^{-3}$ \\
\bottomrule
\toprule
Time history ($th$) & $th = 5$ & $th = 10$ & $th = 20$ \\
\midrule
Test Error & $2.21 \times 10^{-3}$ & $1.96 \times 10^{-3}$ & $1.52 \times 10^{-3}$ \\
\bottomrule
\toprule
Training window & $t \in [10, 100]$ & $t \in [10, 70]$  & $t \in [10, 40]$ \\
\midrule
Test Error & $1.96 \times 10^{-3}$ & $6.42 \times 10^{-3}$ & $5.00 \times 10^{-2}$ \\
\bottomrule
\end{tabular}
\end{table}

We note that the time integrator and autoencoders are the primary workhorse of the transient-CoMLSim model. To this end, table \ref{tab:combined_2-D_swe_ablation} presents the results of varying three important hyperparameters: (a) latent size, (b) time history, and (c) training window, on the quality of the results. We chose 2-D SWE for the ablation study. For each set of runs where a given hyperparameter is varied, the default values for other hyperparameters are as follows: latent size $l=8$, time history $th=10$, and training window $t \in [10, 100]$.

In the ablation study on latent size, we observe that while the test error decreases slightly with increasing latent size, the sensitivity of the model performance is not drastic for the SWE case. This suggests that the optimal latent size may vary depending on the specific application. In this study, all three tested sizes -- 8, 16, and 24 -- performed similarly well. From a model reduction perspective, a larger latent size may introduce noise into the embedding, while a too-small latent size might overlook important large-scale features. Hence, choosing the right latent size requires a balance to capture critical features without adding unnecessary complexity and is very specific to the dynamics of a dataset. For time history, table \ref{tab:combined_2-D_swe_ablation} shows a clear trend: as the time history included in the training process increases, the model accuracy improves. A longer time history provides the model with more temporal information of the history, enabling it to capture the dynamics of the system more effectively. By understanding the evolution of the system over an extended period of time history, the model can better anticipate future states, leading to improved predictions. Conversely, when the training time window is reduced from $t \in [10, 100]$ to $t \in [10, 70]$ to $t \in [10, 40]$, the prediction accuracy of the model deteriorates. This decline occurs because the shorter time window limits the amount of temporal information available, making it harder for the model to accurately capture the underlying patterns and predict future behavior.

\section{Conclusions}\label{sec:conclusions}

In this work, we have proposed a domain decompisition-based deep learning framework to accurately model transient and nonlinear PDEs, building upon its predecessor CoMLSim\cite{ranade2022composable}. Hence, we name the framework transient-CoMLSim. The transient-CoMLSim framework utilizes a CNN-based autoencoder architecture to obtain latent embeddings of the solution and condition variables. Unlike a lot of existing physics-ML frameworks, the autoencoder architecture operates on subdomains instead of the entirety of the computational domain. Each of these subdomains is a collection of computational grid points, e.g., a subdomain of dimension $8\times8\times8$, comprises of 512 computational grid points. Thereafter, an autoregressive model composed of fully connected layers process these embeddings to obtain latent embeddings for the future timesteps. Both the autoencoder and autoregressive model are trained separately and are used in conjunction during the inference. Deriving inspiration from the success of previous works \cite{bengio2015scheduled, ghulenlp}, we employ curriculum learning (CL) technique while training the autoregressive model that gradually allows the model to use its own predictions as an input, resulting in increased robustness and accuracy for long time horizon unrolling. The framework also allows for imposing Dirichlet and periodic boundary conditions in the latent space.

We tested the transient-CoMLSim on four datasets that exhibit different physical dynamics: the 2-D shallow water equation, 2-D diffusion-reaction, 2-D vortex flow, and 3-D additive manufacturing. The last dataset, 3-D additive manufacturing, is an industrial use case designed to evaluate the transient-CoMLSim’s ability to accurately predict on (a) out-of-distribution domain sizes and (b) time horizons much larger than those seen during training. For the 2-D datasets, we demonstrate that our model outperforms two popular models widely used in physics-based deep learning: U-Net \cite{ronneberger2015u} and the Fourier Neural Operator (FNO) \cite{li2020fourier}, in both temporal interpolation and extrapolation tasks.

For the 3-D additive manufacturing dataset, we encountered memory limitations when attempting to fit FNO or U-Net on a single GPU (32 GB capacity), due to the 3-D nature of the problem and the large number of rollout steps required during training. However, the transient-CoMLSim’s ability to operate (a) in the latent space and (b) on local subdomains allows us to successfully train and infer on the 3-D additive manufacturing dataset. Transient-CoMLSim accurately predicts the temperature distribution resulting from a moving laser source for time horizons $3-4$ times longer than those used during training (see figures \ref{fig:additive_mp_plots} and \ref{fig:nrmse_interp}d). Additionally, due to its capability to operate and train on subdomains, transient-CoMLSim can be trained on simulations of smaller computational domain sizes and then used infer on simulations with larger computational domain sizes (see figures \ref{fig:additive_contours} and \ref{fig:additive_mp_plots}).

We also performed ablation studies on the time integrator and autoencoder network for the 2-D SWE dataset, varying key parameters such as the latent embedding size, the time history (\(th\)) provided to the \(\mathbf{TI}\) network during training, and the time window seen by the model during training. Our findings indicate that increasing the time history (\(th\)) leads to an improvement in the model accuracy, as it allows the network to capture more temporal dependencies and time history information. Similarly, expanding the training time window improves model accuracy, suggesting that a bigger temporal window enables the model to better understand and predict the evolution of complex systems. These results emphasize the importance of carefully selecting $\mathbf{TI}$ hyperparameters to optimize model performance. For the 2-D SWE case, we found that the model's performance was not particularly sensitive to the latent size. However, this sensitivity could vary with different datasets.

In future work, the current framework can be enhanced by integrating Runge-Kutta methods for more accurate time-stepping and incorporating Neumann boundary conditions to broaden the scope of applicable problems. We also plan to explore the possibility of training the time integrator and autoencoder in conjunction, thereby creating a more tightly coupled framework that could improve overall prediction accuracy and stability. Additionally, extending this approach to unstructured grids is a natural progression, allowing for the modeling of more complex geometries and real-world scenarios. These advancements will both increase the robustness of the framework and expand its applicability to a wider range of challenging computational problems.

% \begin{acknowledgments}
% We wish to acknowledge the support of the author community in using
% REV\TeX{}, offering suggestions and encouragement, testing new versions,
% \dots.
% \end{acknowledgments}

\section*{Data Availability Statement}
The data that support the findings of this study are available from the corresponding author upon reasonable request.

\bibliography{aipsamp}

\clearpage

\appendix
\section{Animations for the four datasets} \label{sec:appendix}

\begin{figure}[ht]
\centering
\includegraphics[trim={0.0cm, 2cm, 0.0cm, 0cm},clip=true,width=1\linewidth]{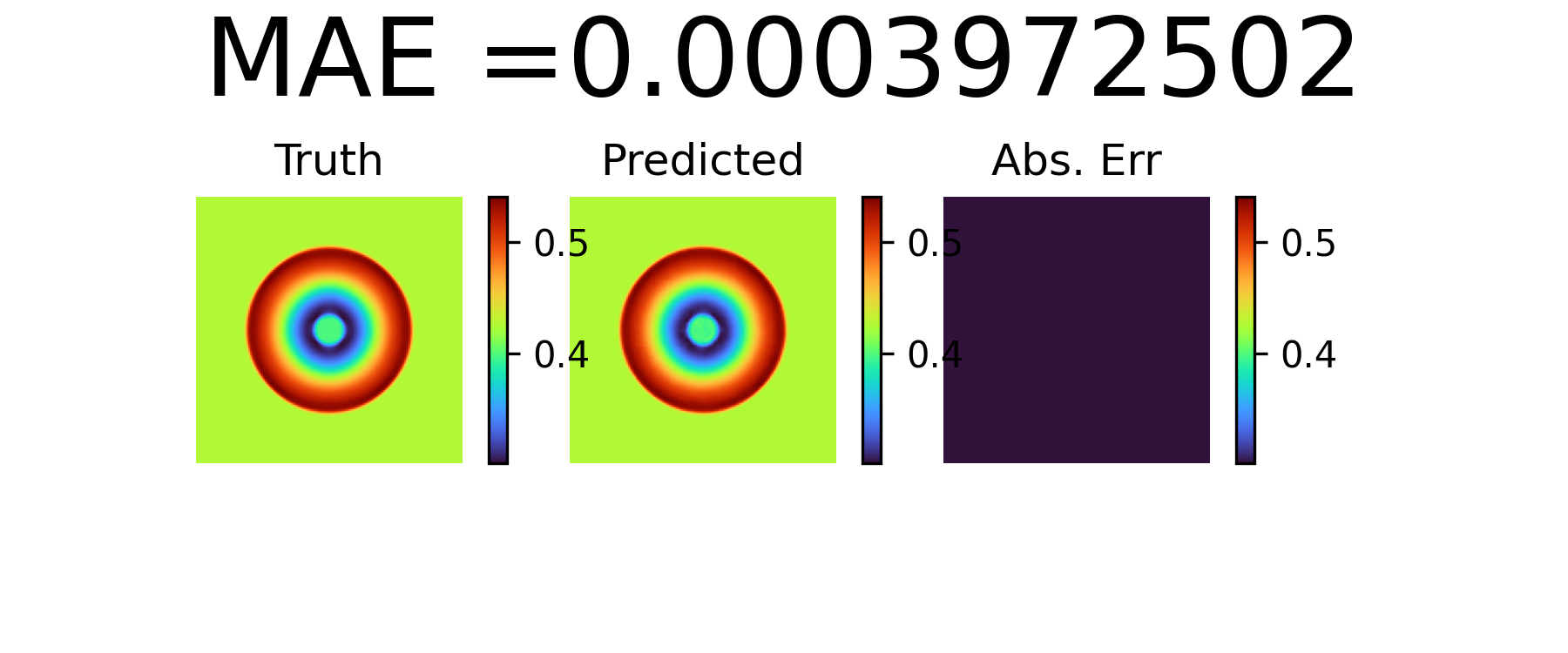}
\caption{SWE animation - random sample 1 (multimedia available online).}
\label{fig:output_swe1}
\end{figure}

\begin{figure}[ht]
\centering
\includegraphics[trim={0.0cm, 2cm, 0.0cm, 0cm},clip=true,width=1\linewidth]{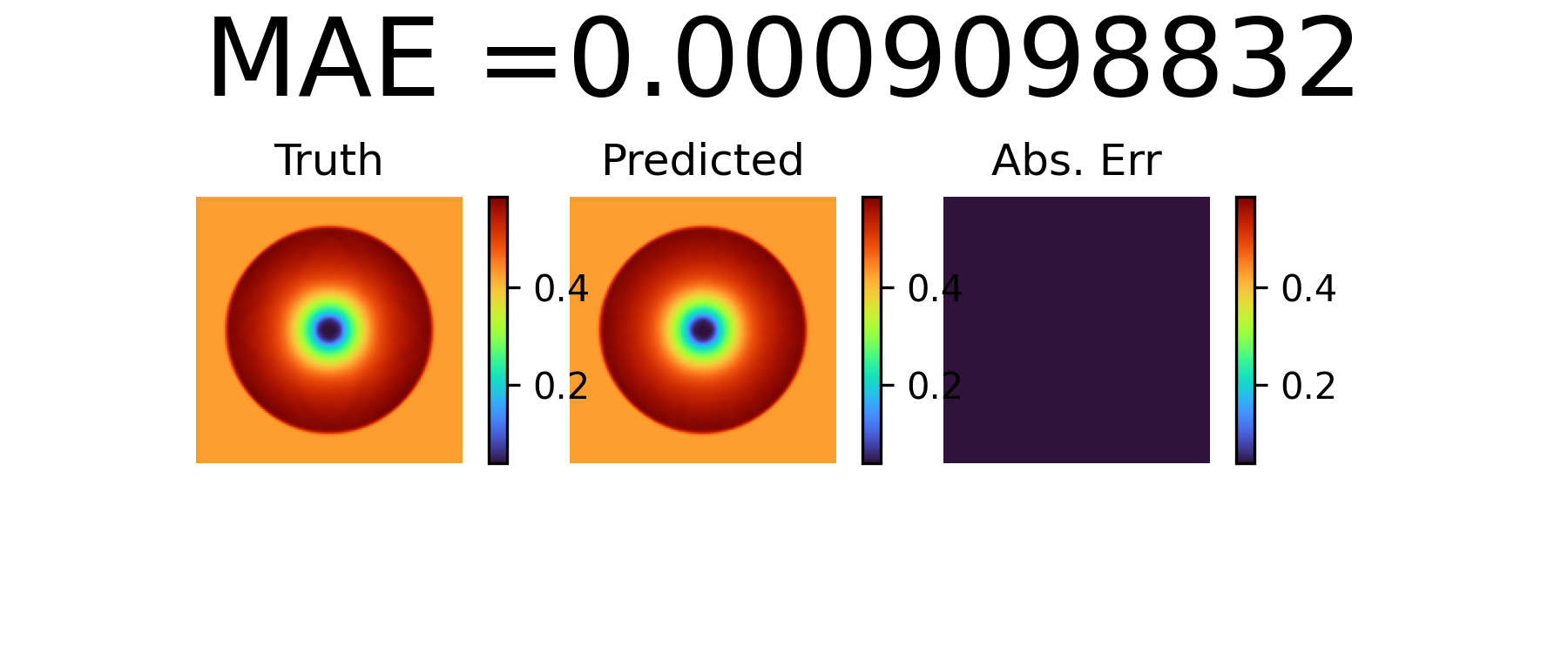}
\caption{SWE animation - random sample 3 (multimedia available online).}
\label{fig:output_swe3}
\end{figure}

\begin{figure}[ht]
\centering
\includegraphics[trim={0.0cm, 2cm, 0.0cm, 0cm},clip=true,width=1\linewidth]{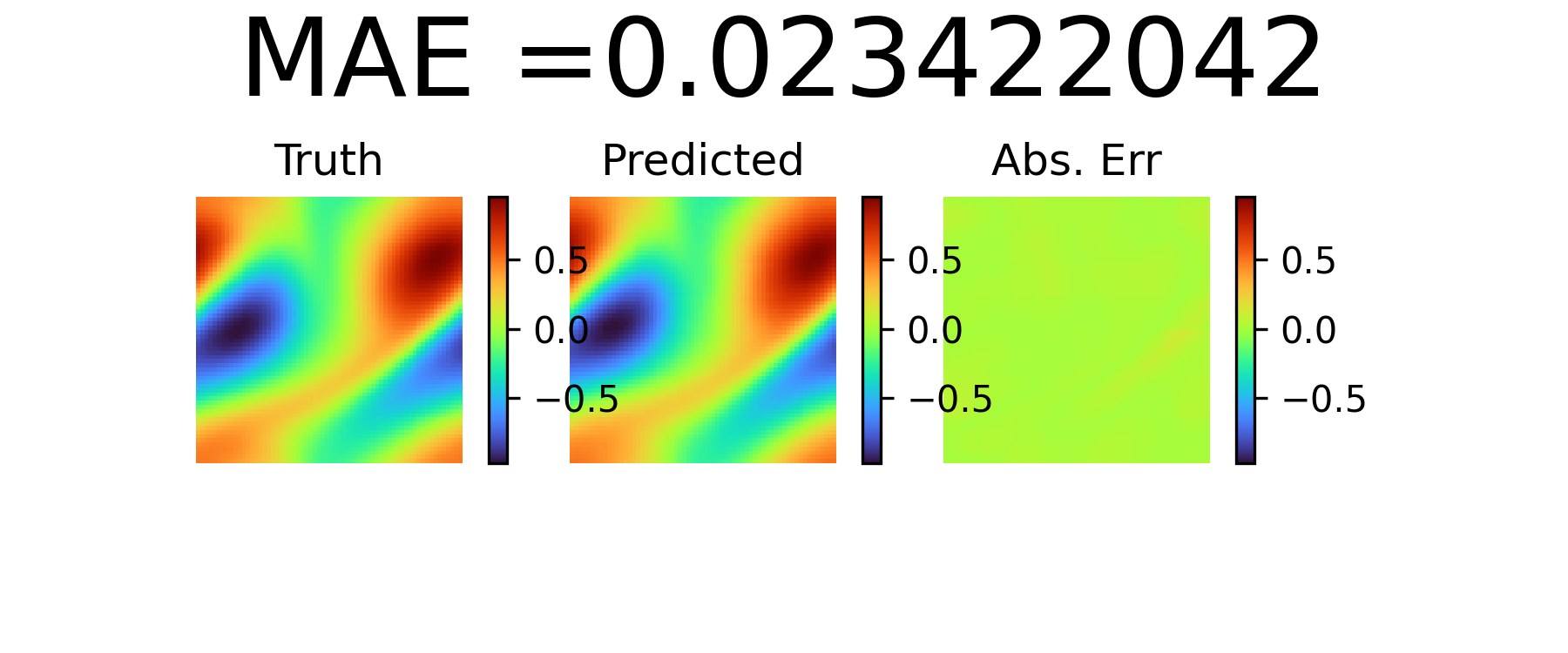}
\caption{Vortex flow animation - random sample 1 (multimedia available online).}
\label{fig:output_fno1}
\end{figure}

\begin{figure}[ht]
\centering
\includegraphics[trim={0.0cm, 2cm, 0.0cm, 0cm},clip=true,width=1\linewidth]{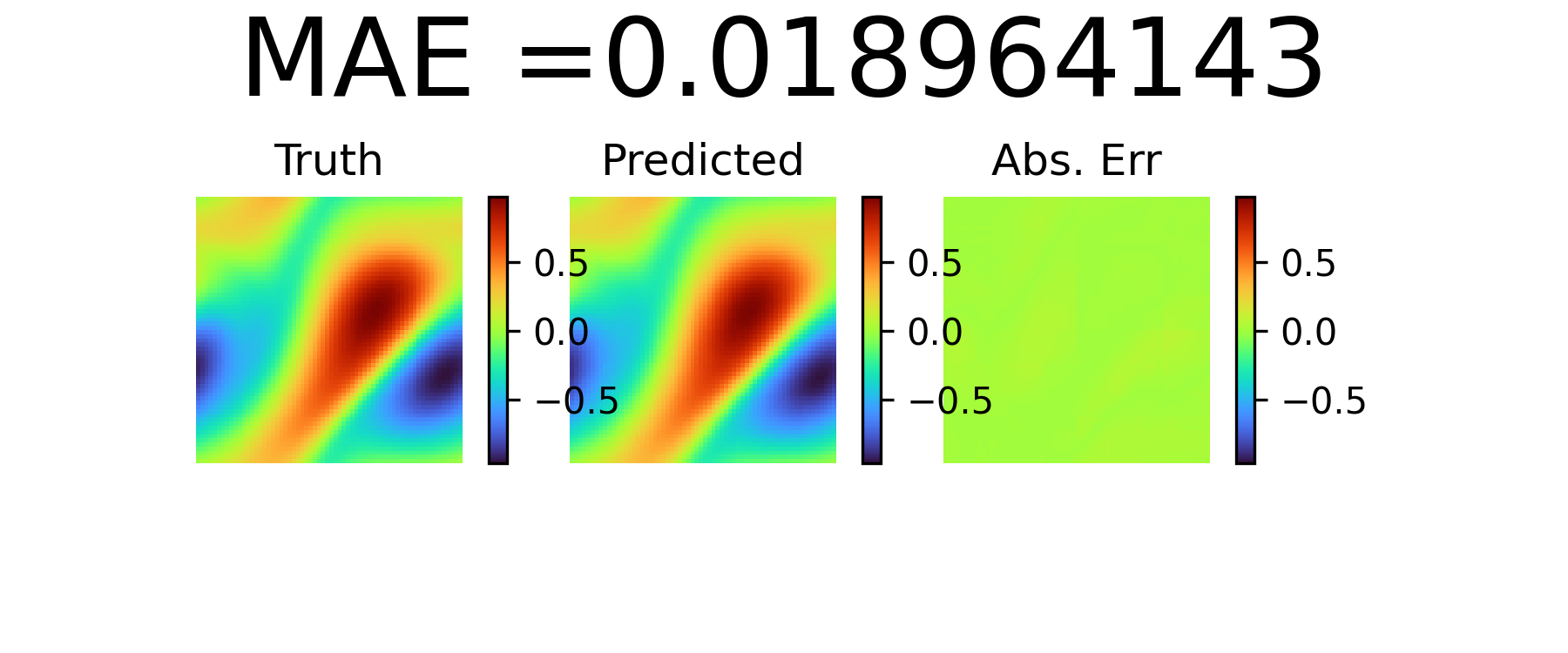}
\caption{Vortex flow animation - random sample 3 (multimedia available online).}
\label{fig:output_fno3}
\end{figure}

\begin{figure}[ht]
\centering
\includegraphics[trim={0.0cm, 2cm, 0.0cm, 0cm},clip=true,width=1\linewidth]{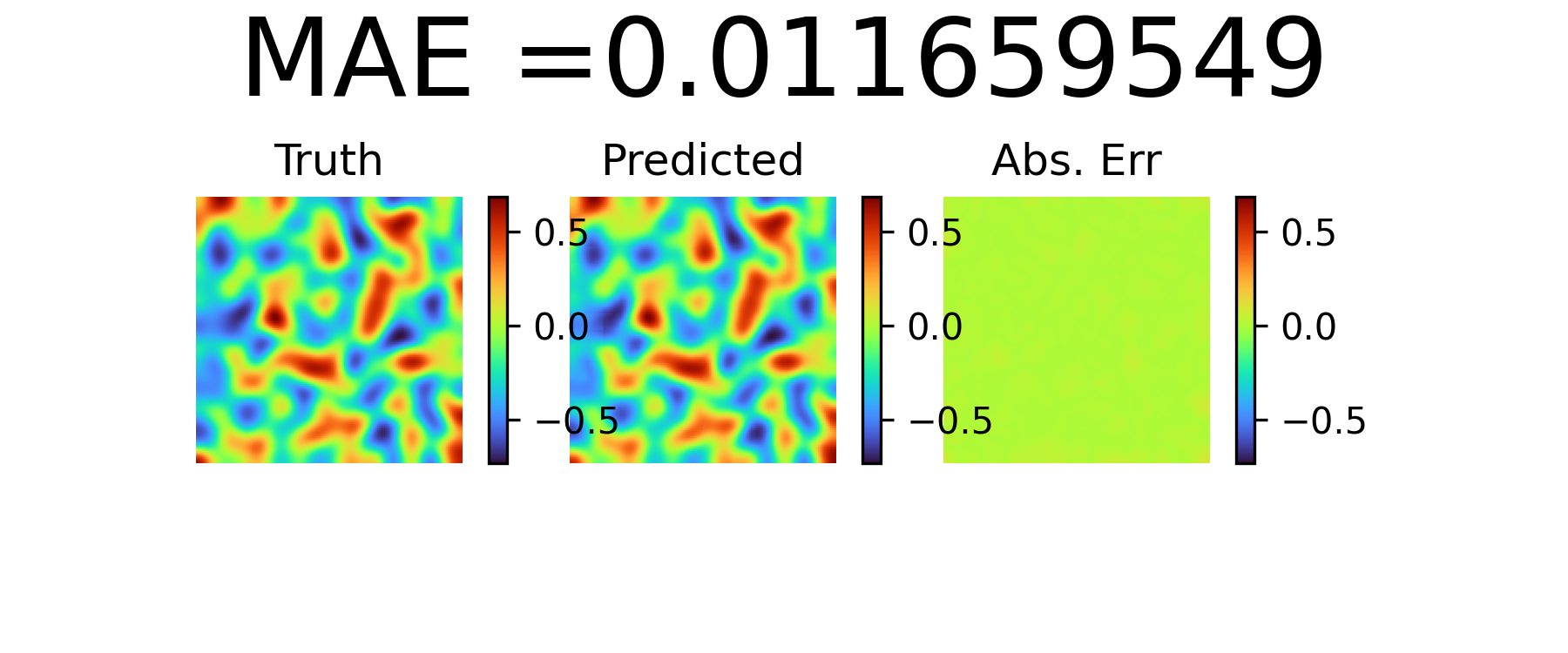}
\caption{Diffusion reaction animation - random sample 1 - variable $u$ (multimedia available online).}
\label{fig:output_diff1}
\end{figure}

\begin{figure}[ht]
\centering
\includegraphics[trim={0.0cm, 2cm, 0.0cm, 0cm},clip=true,width=1\linewidth]{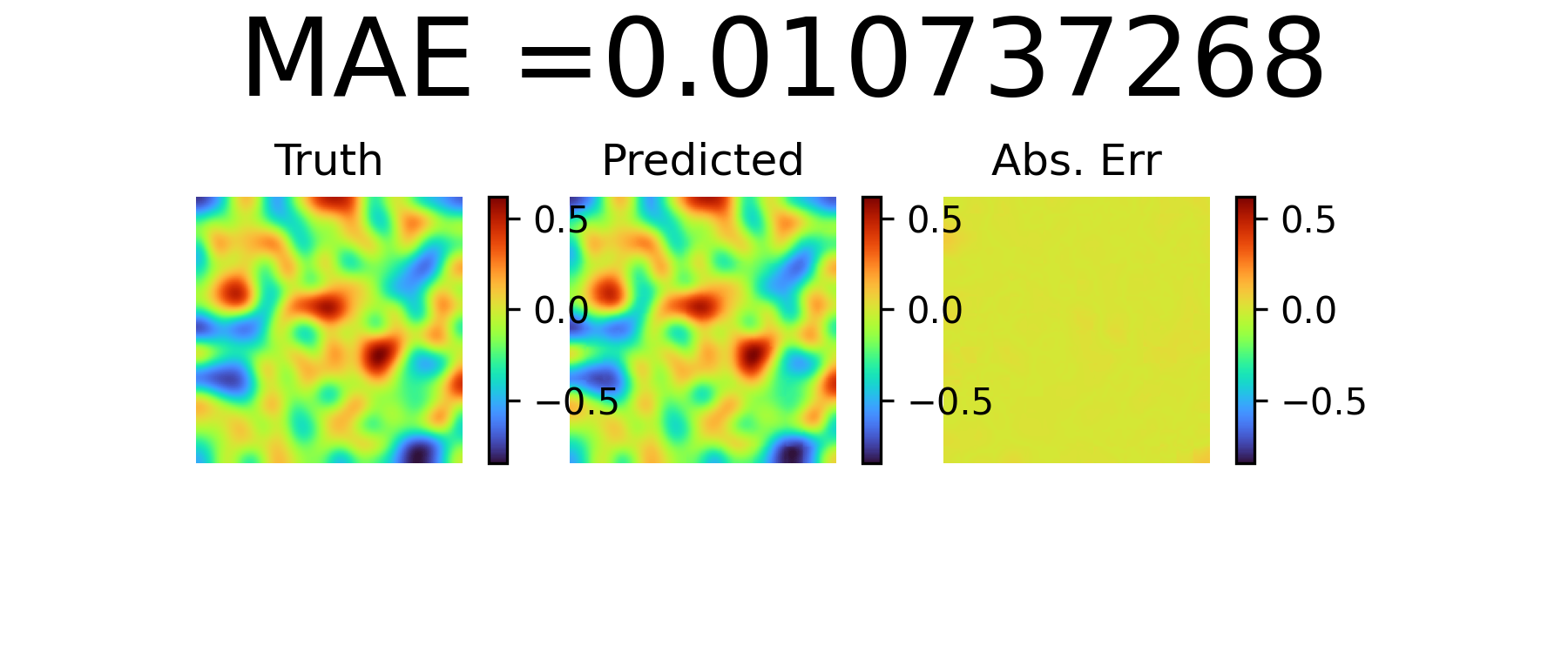}
\caption{Diffusion reaction animation - random sample 3 - variable $v$ (multimedia available online).}
\label{fig:output_diff3}
\end{figure}

\begin{figure}[ht]
\centering
\includegraphics[trim={0.0cm, 0cm, 0.0cm, 0cm},clip=true,width=1\linewidth]{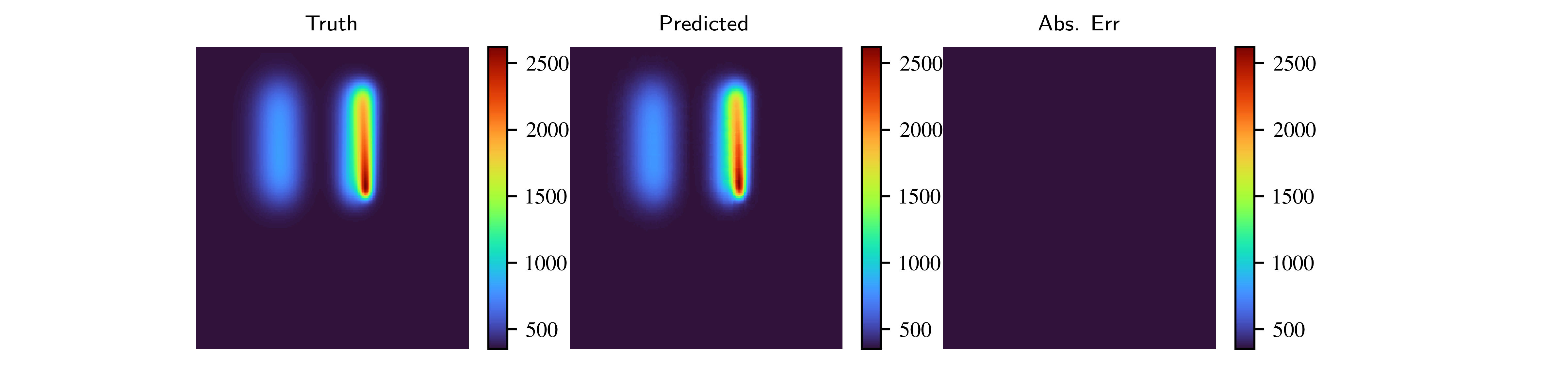}
\caption{AM1 animation (multimedia available online).}
\label{fig:am1}
\end{figure}

\begin{figure}[ht]
\centering
\includegraphics[trim={0.0cm, 0cm, 0.0cm, 0cm},clip=true,width=1\linewidth]{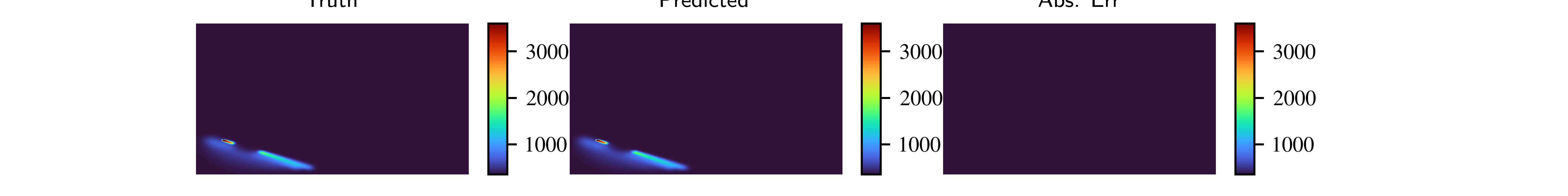}
\caption{AM5 animation (multimedia available online).}
\label{fig:am5}
\end{figure}

Figures \ref{fig:output_swe1} \ref{fig:output_swe3} present animations of the results for three randomly selected test samples from the 2-D SWE dataset. Figures \ref{fig:output_fno1}--\ref{fig:output_fno3} display animations for three randomly selected test samples from the 2-D vortex flow dataset. Figures \ref{fig:output_diff1}--\ref{fig:output_diff3} show the evolution of both the $u$ and $v$ variables for three randomly selected samples from the 2-D diffusion-reaction dataset. Finally, figures \ref{fig:am1}-\ref{fig:am5} present animations for the $\mathrm{AM}1$--$\mathrm{AM5}$ 3-D additive manufacturing test datasets.

\begin{figure}[ht]
\centering
\includegraphics[trim={0.0cm, 2cm, 0.0cm, 0cm},clip=true,width=1\linewidth]{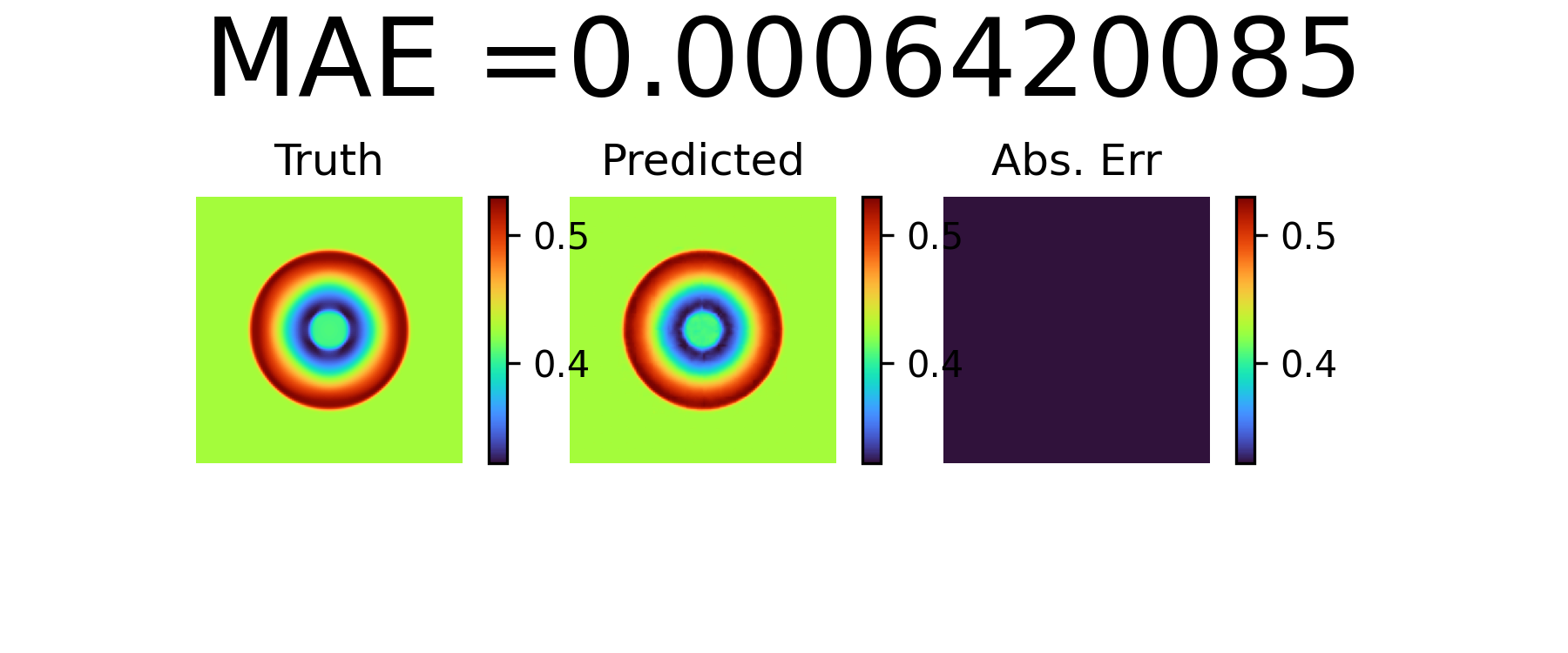}
\caption{SWE animation - random sample 2 (multimedia available online).}
\end{figure}

\begin{figure}[ht]
\centering
\includegraphics[trim={0.0cm, 2cm, 0.0cm, 0cm},clip=true,width=1\linewidth]{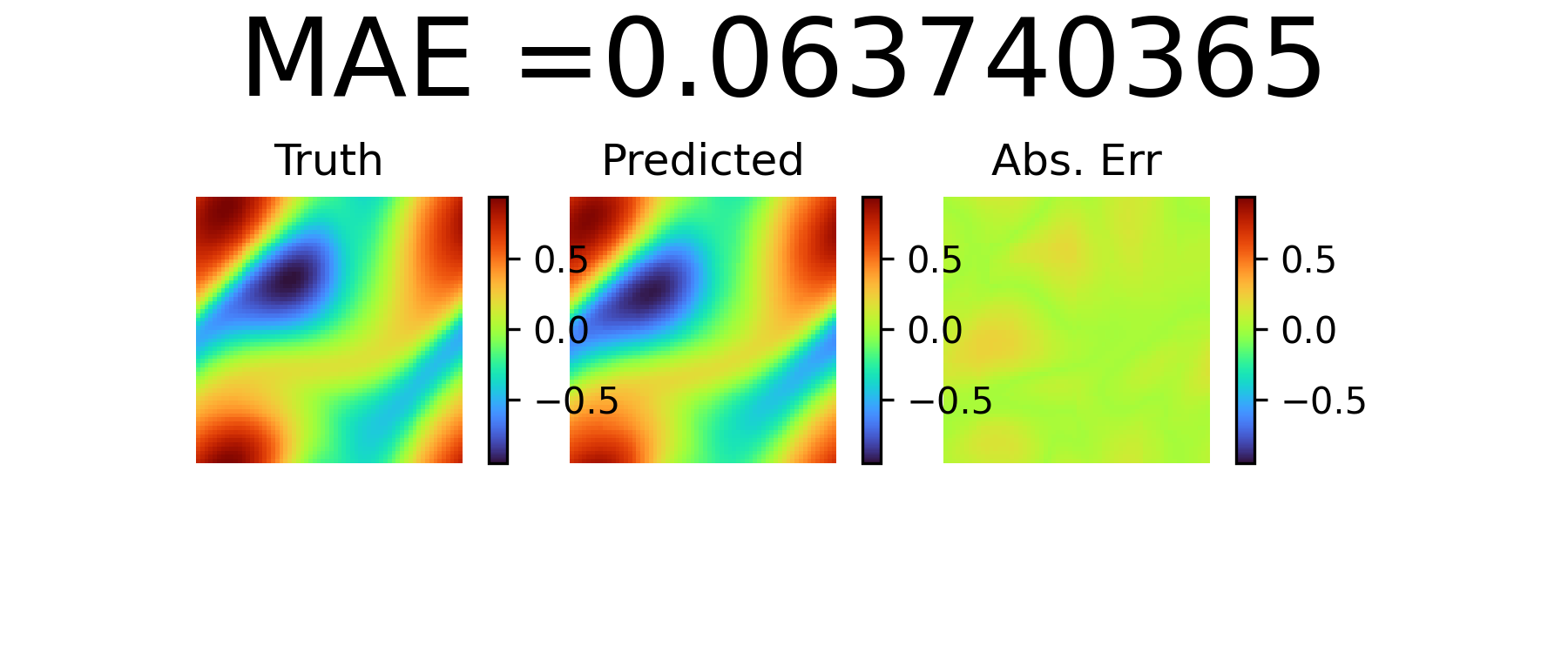}
\caption{Vortex flow animation - random sample 2 (multimedia available online).}
\end{figure}

\begin{figure}[ht]
\centering
\includegraphics[trim={0.0cm, 2cm, 0.0cm, 0cm},clip=true,width=1\linewidth]{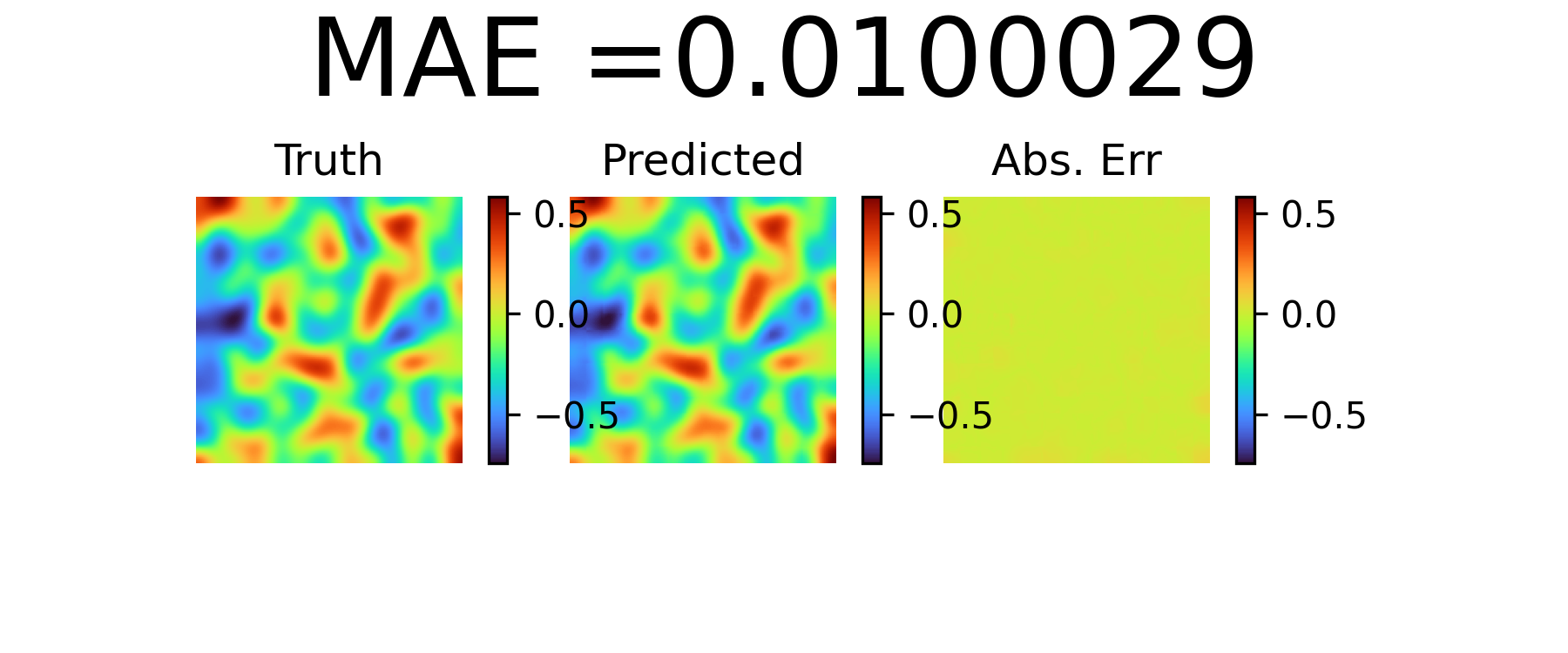}
\caption{Diffusion reaction animation - random sample 1 - variable $v$ (multimedia available online).}
\end{figure}

\begin{figure}[ht]
\centering
\includegraphics[trim={0.0cm, 2cm, 0.0cm, 0cm},clip=true,width=1\linewidth]{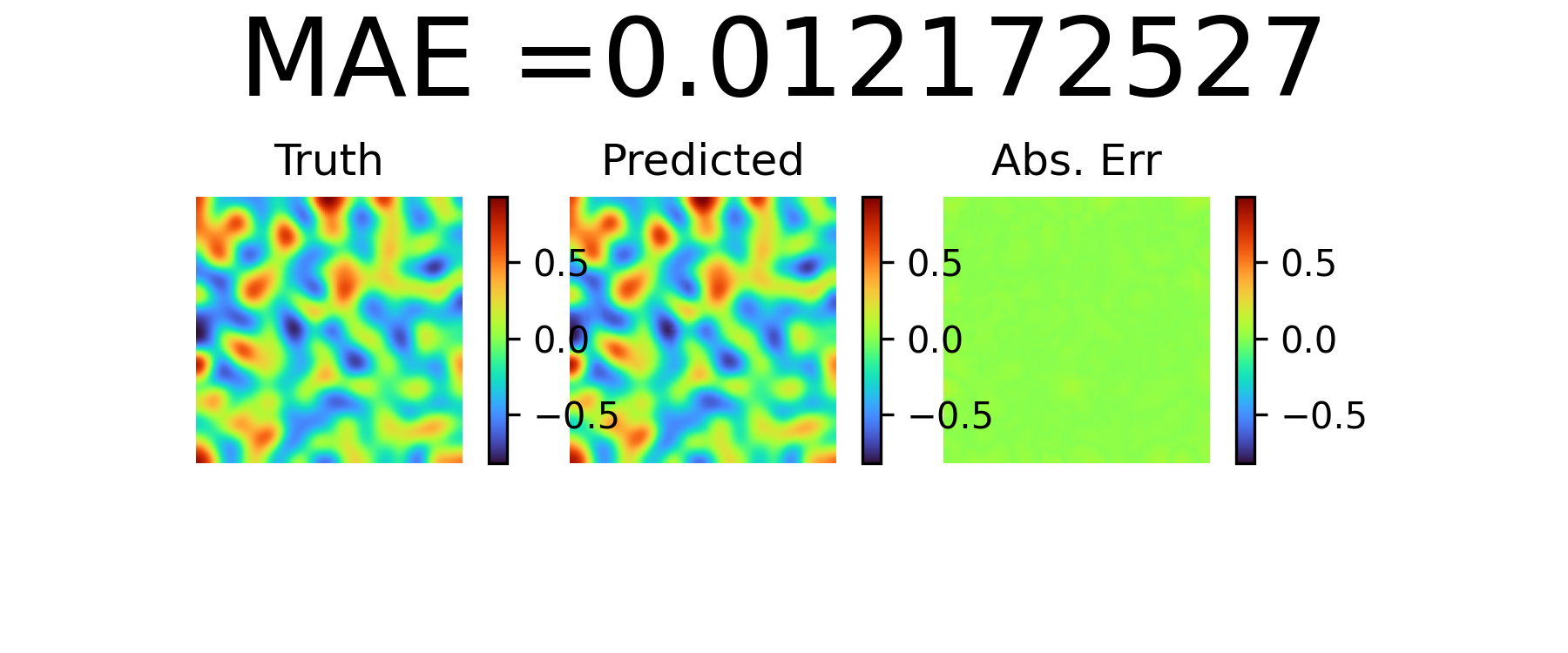}
\caption{Diffusion reaction animation - random sample 2 - variable $u$ (multimedia available online).}
\end{figure}

\begin{figure}[ht]
\centering
\includegraphics[trim={0.0cm, 2cm, 0.0cm, 0cm},clip=true,width=1\linewidth]{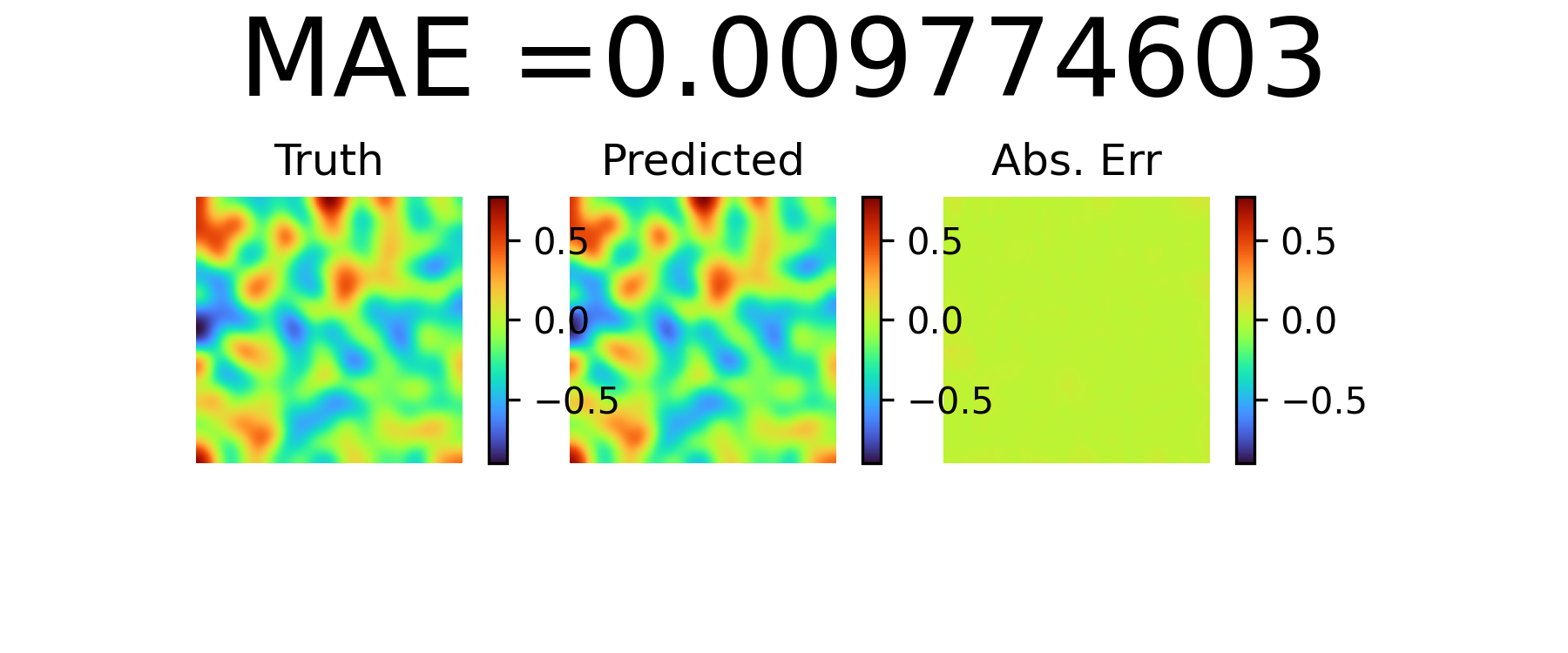}
\caption{Diffusion reaction animation - random sample 2 - variable $v$ (multimedia available online).}
\end{figure}

\begin{figure}[ht]
\centering
\includegraphics[trim={0.0cm, 2cm, 0.0cm, 0cm},clip=true,width=1\linewidth]{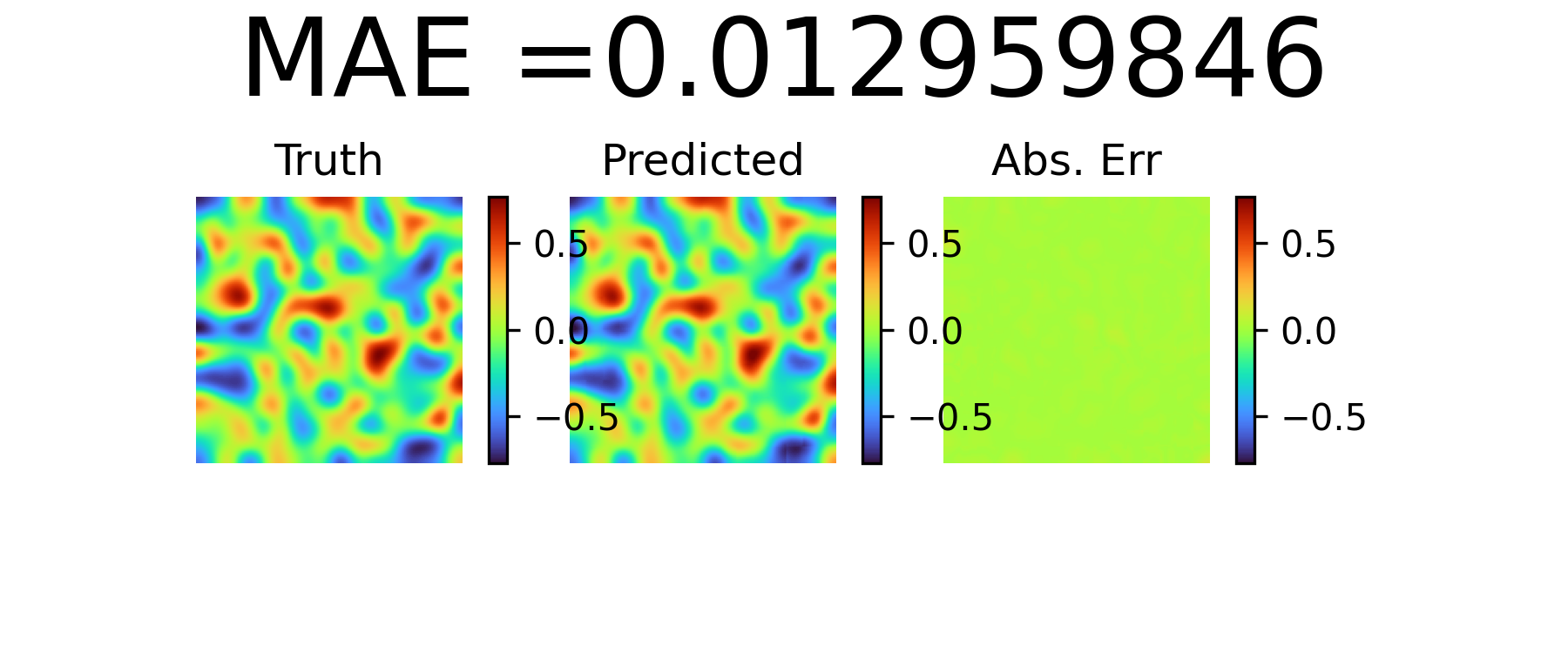}
\caption{Diffusion reaction animation - random sample 3 - variable $u$ (multimedia available online).}
\end{figure}

\begin{figure}[ht]
\centering
\includegraphics[trim={0.0cm, 0cm, 0.0cm, 0cm},clip=true,width=1\linewidth]{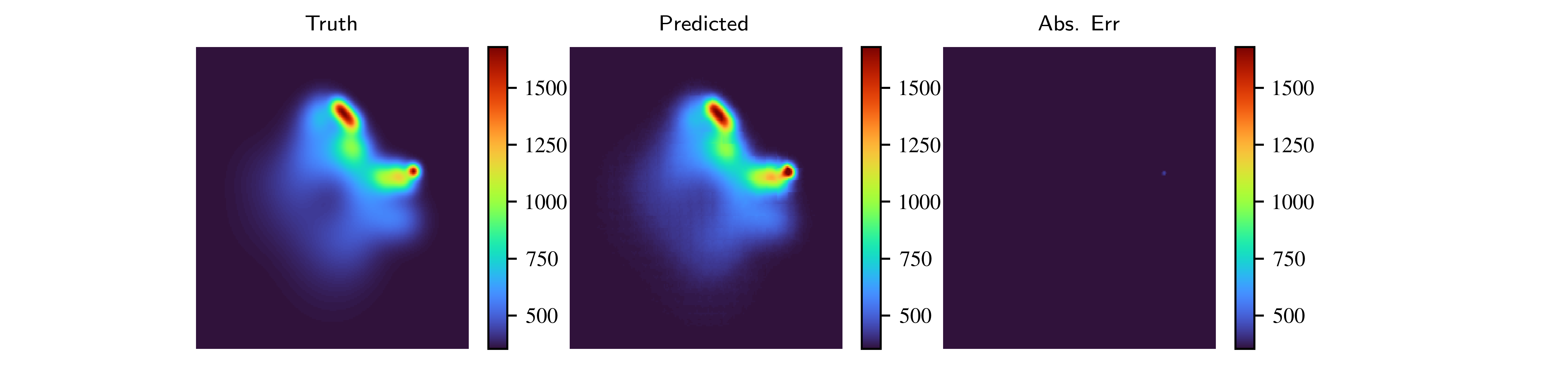}\caption{AM2 animation (multimedia available online).}
\end{figure}

\begin{figure}[ht]
\centering
\includegraphics[trim={0.0cm, 0cm, 0.0cm, 0cm},clip=true,width=1\linewidth]{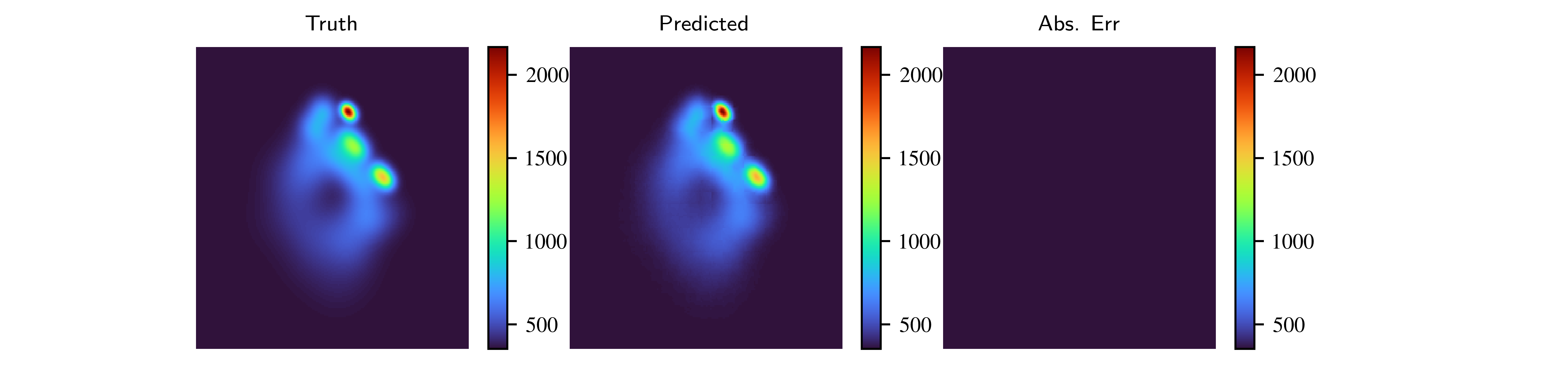}\caption{AM3 animation (multimedia available online).}
\end{figure}

\begin{figure}[ht]
\centering
\includegraphics[trim={0.0cm, 0cm, 0.0cm, 0cm},clip=true,width=1\linewidth]{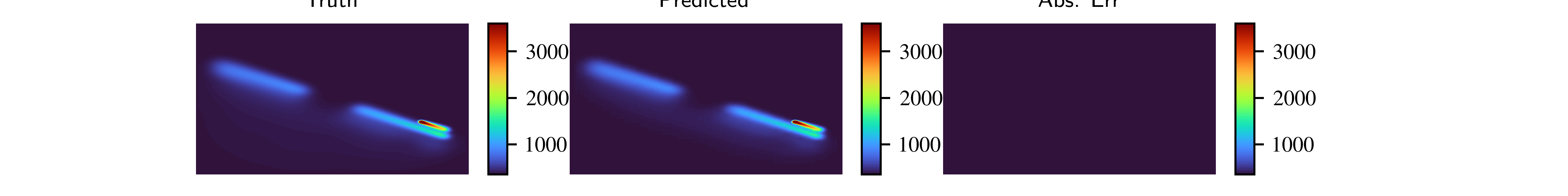}\caption{AM4 animation (multimedia available online).}
\end{figure}

\end{document}